\documentclass[10pt,twocolumn,letterpaper]{article}

\usepackage{cvpr}              
\usepackage{tabularx}
\usepackage{optidef}
\usepackage{multirow}
\usepackage{bm}
\usepackage{tabularx}
\usepackage{pgfplots}
\usepgfplotslibrary{statistics}
\usepackage{pgf,tikz}
\usepackage{mathtools}
\usepackage{amsmath}
\usepackage{algorithm}
\usepackage{algpseudocode}
\usepackage{float}
\usepackage[accsupp]{axessibility}

\definecolor{cPLOTours}{RGB}{214,113,176}
\definecolor{cPLOTdpfm}{RGB}{80,150,80}
\definecolor{cPLOTroetzer}{RGB}{165, 124, 27}
\definecolor{cPLOT2}{RGB}{68, 33, 175}
\definecolor{cPLOT5}{RGB}{39, 174, 239}
\definecolor{cPLOT6}{RGB}{179,0,0}

\definecolor{cvprblue}{rgb}{0.21,0.49,0.74}
\usepackage[pagebackref,breaklinks,colorlinks,citecolor=cvprblue]{hyperref}

\def\paperID{11786} 

\def\confName{CVPR}
\def\confYear{2024}

\title{Partial-to-Partial Shape Matching with Geometric Consistency}

\author{Viktoria Ehm$^{1,2}$~~~~~~~~~
Maolin Gao$^{1,2}$~~~~~~~~~
Paul Roetzer$^{3}$~~~~~~~~~
Marvin Eisenberger$^{1,2}$\\
Daniel Cremers$^{1,2}$~~~~~~~~~
Florian Bernard$^{3}$
\\
$^{1}$Technical University of Munich~~~~$^{2}$Munich Center for Machine Learning~~~~$^{3}$University of Bonn}

\begin{document}
\twocolumn[{%
\renewcommand\twocolumn[1][]{#1}%
\maketitle
\begin{center}%
    \centering%
    \captionsetup{type=figure}%
   \includegraphics[width=1\textwidth]{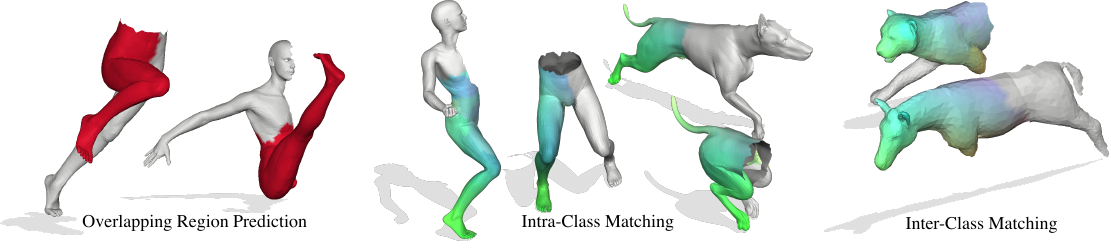}
    \captionof{figure}{We present the first \textbf{geometrically consistent partial-to-partial shape matching} solution.
    Our approach can find correspondences between partial shapes and thus determine the overlapping region between these shapes (left).
    We can find correspondences within shape classes (middle), and across shape classes (right). For the latter, we introduce a new inter-class partial-to-partial matching dataset.
    }
\end{center}%
}]
\begin{abstract}
Finding correspondences between 3D shapes is an important and long-standing problem in computer vision, graphics and beyond.
A prominent challenge are partial-to-partial shape matching settings, which occur when the shapes to match
are only observed incompletely (e.g.~from 3D scanning). Although partial-to-partial matching is a highly relevant setting in practice, it is rarely explored. Our work bridges the gap between existing (rather artificial) 3D full shape matching and partial-to-partial real-world settings by exploiting geometric consistency as a strong constraint. We demonstrate that it is indeed possible to solve this challenging problem in a variety of settings. For the first time, we achieve geometric consistency for partial-to-partial matching, which is realized by a novel integer non-linear program formalism building on triangle product spaces, along with a new pruning algorithm based on linear integer programming.  
Further, we generate a new inter-class dataset for partial-to-partial shape-matching.
We show that our method outperforms current SOTA methods on both an established intra-class dataset and our novel inter-class dataset.
{The code of this work is publicly available\footnote{\url{https://github.com/vikiehm/gc-ppsm}}}.
\end{abstract}

\section{Introduction}
\label{sec:intro}
Finding correspondences between 3D shapes is an important topic in computer vision to solve tasks such as shape interpolation or texture transfer.
While there is a lot of research on 3D shape matching, these works predominantly focus on full-to-full shape matching \cite{ovsjanikov2012functional, cao2023unsupervised, eisenberger2020, li2022learning, windheuser2011geometrically, windheuser2011large, Schmidt-et-al-14}, or on partial-to-full shape matching \cite{rodola2017partial, litany2017fully, cao2023unsupervised, attaiki2021dpfm, roetzer2022scalable, ehm2023geometrically}.
Therefore, they are not applicable to the complex problem of matching two partial shapes.
The problem of partial-to-partial matching is difficult because not every part of one shape will necessarily be matched to the other shape and vice versa. 
At the same time, we must encourage specific parts to be matched to avoid the trivial solution of not matching anything.
There is only little research on the topic of partial-to-partial shape matching.

The current SOTA method~\cite{attaiki2021dpfm}
requires annotated data for training, which is often unavailable rendering many applications infeasible.
While the method~\cite{roetzer2022scalable} is
{learning-free and thus does not rely on annotations},
it requires closing holes in the partial shapes to generate a full-to-full shape matching, resulting in 
{poor results for shapes with low overlaps.}
It uses the property of geometric consistency to generate its results.
While many methods also make use of the connectivity of meshes to generate geometrically consistent results for full-to-full~\cite{windheuser2011large, Schmidt-et-al-14, windheuser2011geometrically} or partial-to-full~\cite{ehm2023geometrically} matching, to date, it has not been applied to partial-to-partial matching.
In this work, we incorporate geometric consistency into a partial-to-partial  shape matching setting by formulating and solving a non-linear integer program with a novel pruning algorithm.
Most current SOTA shape matching methods use deep functional maps~\cite{li2022learning, cao2022unsupervised, donati2022deep, cao2023unsupervised, attaiki2022ncp} 
to find correspondences between 3D shapes.
Functional maps are not directly applicable to partial-to-partial shape matching because they cannot predict the overlapping region. 
We integrate deep features from the SOTA feature extractor~\cite{cao2023unsupervised} into our partial-to-partial integer program.
Current partial-to-partial shape matching approaches focus solely on intra-class datasets~\cite{attaiki2021dpfm}. Instead, we present a new inter-class dataset based on SMAL~\cite{Zuffi:CVPR:2017} to expand the scope of partial-to-partial shape matching evaluation.
We summarize our contributions as follows:
\begin{itemize}
    \item We present a \textbf{geometrically consistent partial-to-partial} shape matching formalism, that fuses \textbf{SOTA deep features} with a \textbf{non-linear integer programming approach} to combine the best of both worlds.
    \item We introduce a \textbf{pruned search algorithm}
    to solve the integer program in a reasonable time.
    \item We provide a new \textbf{inter-class} partial-to-partial dataset based on the SMAL dataset.
    \item We show that our approach \textbf{outperforms} SOTA supervised deep learning and combinatorial optimization algorithms in both intra-class and inter-class settings.
\end{itemize}

\section{Related Work}
While the entire field of 3D shape matching is vast, this paper focuses on closely related work. 
For a more comprehensive overview, interested readers are referred to the following 
surveys~\cite{van2011survey, sahilliouglu2020recent}.

\subsection{Integer and Combinatorial Optimization in Shape Matching}
Many shape correspondence formalisms  
involve integer or binary variables, 
for which permutation matrices are a popular representation.
Depending on the objective function (linear or quadratic), a linear assignment problem (LAP) can be solved to global optimality in polynomial time, e.g.~using Bertsekas' Auction algorithm \cite{Bertsekas:1998vt}, while the quadratic assignment problem (QAP) is NP-hard~\cite{Loiola:2ua4FrR7, rendl1994quadratic} and thus several relaxed formulations have been proposed~{\cite{bernard:2018, kushinsky2019sinkhorn, Dym:2017ue,haller2022comparative}}.

Instead of matching vertices to vertices, the authors of \cite{windheuser2011geometrically, windheuser2011large, Schmidt-et-al-14} proposed to match mesh faces to faces by minimizing an elastic energy, which models the bending and stretching of the triangles locally, while constraining the matching to be geometrically consistent~\cite{botsch2010polygon}. 
Geometric consistency in this case means that neighboring triangles are matched to neighboring triangles, which naturally leads to smooth results.
Respective approaches leads to orientation-preserving matchings by design, but they involve solving a difficult instance of a linear integer program (i.e.~its constraints are  not totally unimodular), which limits the size of the problem it can tackle~\cite{wolsey1998integer}. However, in practice, high resolution shapes are often available, and the mandatory down-sampling step unavoidably leads to a loss of valuable 3D information, and hence sub-optimal correspondences. Recently, \cite{roetzer2022scalable} has proposed a tailored linear integer solver to improve the method's scalability in trade of compromising the global optimality. 

Most recently, shape matching problems were also modeled as mixed integer programs \cite{bernard2020mina, gao2023sigma}, which can (often) be solved to global optimality by branch-and-bound methods. Though their potential to be applied to partial shapes matching has been shown in their paper, an adequate quantitative study is not presented.

\subsection{Partial-to-Full Shape Matching}
Unlike matching full shapes, partial shapes often exhibit substantial non-isometric and irregular characteristics, making them a challenging problem to address.
Nevertheless, investigating this problem is essential, given its significant practical relevance.

In~\cite{rodola2017partial} a 
method for matching partial shapes to full template shapes has been introduced. 
It extended the functional map framework, which has achieved great success in full-to-full matching~\cite{ovsjanikov2012functional, donati2020deep,litany2017deep}. 
This partial functional maps framework was 
extended to multiple partial  shapes~\cite{Litany2016}, where every partial shape can be further affected by potential overlaps with other parts, the possibility of missing or redundant parts, and the presence of clutter. Clutter is also addressed in~\cite{Cosmo2016}.
Recent deep learning methods use a functional map layer~\cite{attaiki2021dpfm,cao2023unsupervised} to determine correspondences between a partial and a template shape.
An inherent limitation shared by many methods based on functional maps is to get desired structural properties, like bijectivity and smoothness, as observed in previous studies~\cite{vestner2017pmf, ren2018continuous}.
{In~\cite{roetzer2022scalable} a scalable solver was introduced for the formalism of \cite{windheuser2011geometrically, windheuser2011large, Schmidt-et-al-14}, which ensures smoothness utilizing geometric consistency.}
Respective works are also applicable to partial-to-full shape matching (by closing the holes of the partial shapes), which however increases the difficulty to solve this problem. 
In~\cite{ehm2023geometrically}, these works were extended such that geometrically consistent partial-to-full matchings can be produced without requiring closing holes beforehand. 
While this work is closest to our approach, it is not applicable to partial-to-partial shape matching as the authors expect that the partial shape is  completely matched to (parts of) the full shape.

\subsection{Partial-to-Partial Shape Matching}
Even though partial-to-partial shape matching has high practical relevance, there is still few research about it. \citet{litany2017fully} provide a theoretical concept for partial-to-partial shape matching, but no practical analysis. While~\cite{bensaid2023partial} investigates overlapping regions for pairs of partial shapes, it does not provide a method for shape matching. 
DPFM~\cite{attaiki2021dpfm} and Sm-comb~\cite{roetzer2022scalable} are the only approaches enabling partial-to-partial shape matching enabled by deep learning (DPFM) and combinatorial optimization (Sm-comb). 
While DPFM needs labels for training, these are often not accessible in real-world examples. 
Sm-comb does not need labels, however, it requires the closing of holes in shapes, as it is only applicable to water-tight shapes.
Instead, we propose a novel partial-to-partial shape matching formalism without requiring 3D ground truth labels and closing holes.

\begin{figure*}
\centering
\includegraphics[width=1\textwidth]{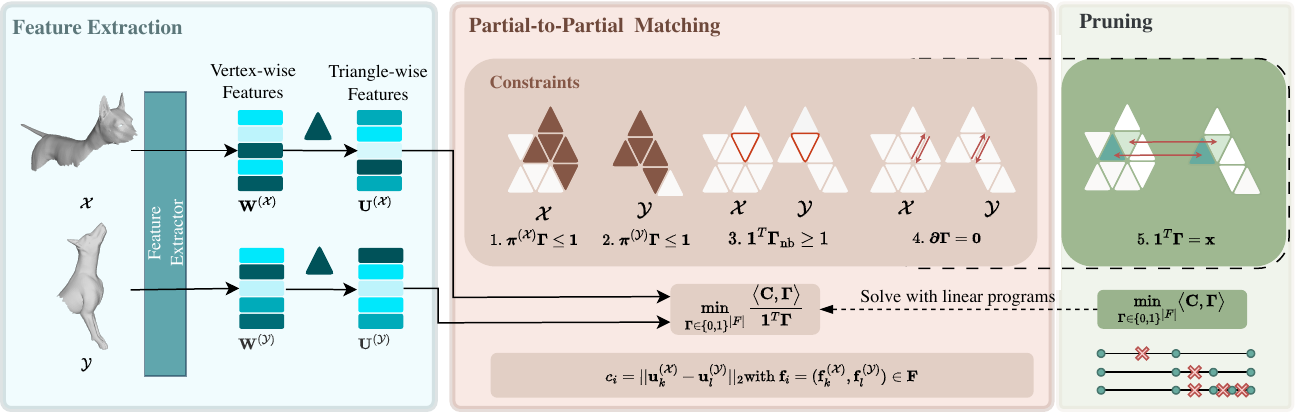}
    \caption{Overview over our \textbf{matching pipeline}: The 3D meshes are fed into a feature extractor that returns vertex-wise features. 
    From these vertex-wise features, we generate triangle-wise features to make them accessible for our algorithm.
    With these features, we define an integer program
    that ensures that every triangle is matched at most once for both shapes, that at least 
    one non-boundary triangle is matched (per shape), and that 
    neighboring relationships for inner triangles are fulfilled. 
    We tackle this non-linear integer program by solving a
    subset of integer linear programs (ILPs) for a specific number of matchings. 
    }
    \label{fig:overview}
\end{figure*}
\section{{Partial-to-Partial Shape Matching}}
Our primary innovation centers around a novel integer programming framework designed for achieving geometrically consistent partial-to-partial shape matching.
We show an overview of our method in Figure~\ref{fig:overview}.
\subsection{Notation}
\paragraph{Shapes.}
We define an oriented manifold shape as 
$\mathcal{X}=\bigl(\mathbf{V}^{(\mathcal{X})},\mathbf{F}^{(\mathcal{X})}\bigr)$, which consists of vertices $\mathbf{V}^{(\mathcal{X})}:=\bigl\{\mathbf{v}^{(\mathcal{X})}_i\in\mathbb{R}^3|1\leq i \leq |\mathbf{V}^{(\mathcal{X})}|\bigr\}$
and triangles $\mathbf{F}^{(\mathcal{X})}\subset\mathbf{V}^{(\mathcal{X})}\times\mathbf{V}^{(\mathcal{X})}\times\mathbf{V}^{(\mathcal{X})}$.
{Moreover, $\mathbf{E}^{(\mathcal{X})}\subset\mathbf{V}^{(\mathcal{X})}\times\mathbf{V}^{(\mathcal{X})}$ are edges induced by triangles.}
The set of edges is divided in boundary $\mathbf{E}^{(\mathcal{X})}_b \subset \mathbf{E}^{(\mathcal{X})}$ and non-boundary $\mathbf{E}^{(\mathcal{X})}_{nb} \subset \mathbf{E}^{(\mathcal{X})}$ edges.

\paragraph{Degenerate Triangles.}
To allow shrinking and expanding in our matching, which is necessary for different discretization and non-isometric shape pairs, we allow to match triangles from one shape not only to triangles from the other shape, but also to vertices and edges from the other shape. 
For this reason, we define a degenerate triangle as either a vertex $\mathbf{v} \in \mathbf{V}^{(\mathcal{X})}$ or an edge $\mathbf{e} \in \mathbf{E}^{(\mathcal{X})}$ and collect these in the set of extended triangles which we define as $\mathbf{\bar{F}}^{(\mathcal{X})}:=\mathbf{V}^{(\mathcal{X})}\cup\mathbf{E}^{(\mathcal{X})}\cup\mathbf{F}^{(\mathcal{X})}$.

\paragraph{Inner Triangles.}
We utilize the property that a non-boundary edge $\mathbf{e} \in \mathbf{E}_{nb}^{(\mathcal{X})}$ is common to two adjacent triangles.
In contrast, a boundary edge $\mathbf{e} \in \mathbf{E}_{b}^{(\mathcal{X})}$ is adjacent to only one triangle.
As the shape $\mathcal{X}$ is oriented, non-boundary edges are found in pairs with opposing orientations within adjacent triangles
$(\mathbf{v}^{(\mathcal{X})}_i, \mathbf{v}^{(\mathcal{X})}_j)$ and $(\mathbf{v}^{(\mathcal{X})}_j, \mathbf{v}^{(\mathcal{X})}_i)$.
We use the notation $\mathcal{O}^{(\mathcal{X})}:\mathbf{E}^{(\mathcal{X})}\times \mathbf{F}^{(\mathcal{X})}\to \{-1,0,1\}$ to represent the orientation operator between an edge $\mathbf{e}=(\mathbf{v}_1',\mathbf{v}_2')$ and a triangle $\mathbf{f}=(\mathbf{v}_1,\mathbf{v}_2,\mathbf{v}_3)$.
The orientation operator represents the orientation of an edge $\mathbf{e}$ in dependence of a triangle $\mathbf{f}$, and whether this edge is adjacent to the triangle (refer to \cite{windheuser2011geometrically, windheuser2011large, roetzer2022scalable, ehm2023geometrically} for further details).

\subsection{{Optimization Problem}}
\label{sec:p2p-shape-matching}
We present a method to match two oriented and manifold shapes, denoted as $\mathcal{X}$ and $\mathcal{Y}$, where both shapes have boundaries. 
With this approach we extend the full-to-full matching algorithm from~\cite{windheuser2011geometrically, windheuser2011large, Schmidt-et-al-14} and the partial-to-full matching from~\cite{ehm2023geometrically} to a partial-to-partial formulation.
\paragraph{Product Space}
We define product edges $\mathbf{E}:=\bigl(\mathbf{E}^{(\mathcal{X})}\times\mathbf{E}^{(\mathcal{Y})}\bigr)\cup\bigl(\mathbf{E}^{(\mathcal{X})}\times\mathbf{V}^{(\mathcal{Y})}\bigr)\cup\bigl(\mathbf{V}^{(\mathcal{X})}\times\mathbf{E}^{(\mathcal{Y})}\bigr)$ by the combination of an edge with an edge or an edge with a vertex.
Additionally, we define product triangles $\mathbf{F}:=\bigl(\mathbf{\bar{F}}^{(\mathcal{X})}\times\mathbf{{F}}^{(\mathcal{Y})}\bigr)\cup\bigl(\mathbf{F}^{(\mathcal{X})}\times\mathbf{\bar{F}}^{(\mathcal{Y})}\bigr)$.
Every product triangle represents a
matching between shape $\mathcal{X}$ and shape $\mathcal{Y}$.
To encode these product triangles we define a binary vector $\mathbf{\Gamma}\in\{0,1\}^{|\mathbf{F}|}$ {that forms the matching for which we optimize for.}
{The value $\Gamma_i=1$} means that the (degenerate or non-degenerate) triangles $f_j \in \mathbf{\bar{F}}^{(\mathcal{X})}$ and $f_k \in \mathbf{\bar{F}}^{(\mathcal{Y})}$, represented by the product triangle $\mathbf{F}_i = \bigl(f_j, f_k)$, are matched.

\paragraph{Uniqueness Constraints.}
In the context of partial shape matching, triangles can either be matched or not be matched. We denote matched triangles as \emph{overlapping}.
Our optimization approach ensures that each triangle within the partial shapes is matched at most once, emphasizing the uniqueness of correspondences in the matching process.
Following the approach of prior studies~\cite{windheuser2011geometrically, ehm2023geometrically}, we express this condition through the use of projection matrices. $\bm{\pi}^{(\mathcal{X})}\in\{0,1\}^{|\mathbf{F}^{(\mathcal{X})}|\times|\mathbf{F}|}$ and $\bm{\pi}^{(\mathcal{Y})}\in\{0,1\}^{|\mathbf{F}^{(\mathcal{Y})}|\times|\mathbf{F}|}$.
For every combination between a triangle $\mathbf{f}_i^{(\mathcal{X})}$ in $\mathcal{X}$ and a product triangle $\mathbf{f}_j$, the projection entry is defined as
\begin{equation}
\bm{\pi}^{(\mathcal{X})}_{i,j}:=\begin{cases}
    1 & \text{if}\quad \mathbf{f}_j \text{ contains } \mathbf{f}_i^{(\mathcal{X})} \\
    0 &\text{else.}
    \end{cases}
\end{equation}
The definition of $\bm{\pi}^{(\mathcal{Y})}\in\{0,1\}^{|\mathbf{F}^{(\mathcal{Y})}|\times|\mathbf{F}|}$ follows similarly.
Whereas in a full matching every triangle must be matched exactly once, i.e., $\bm{\pi}^{(\mathcal{X})} \mathbf{\Gamma}= \mathbf{1}$, in the partial-to-partial matching scenario we merely impose that each triangle is matched {\em at most} once.
The uniqueness constraints are defined as
\begin{equation}
\label{eq:op-proj}
    \bm{\pi}^{(\mathcal{X})} \mathbf{\Gamma}\leq \mathbf{1},\quad
    \bm{\pi}^{(\mathcal{Y})} \mathbf{\Gamma} \leq \mathbf{1},
\end{equation}
where $\mathbf{1}$ is a vector of ones.
\paragraph{Minimum Triangle Matching.}
When minimizing the costs ({which} {are non-negative and} will be introduced in the paragraph \textbf{Cost Function} below) without considering additional constraints, the best solution would be that no triangles are matched.
To avoid this degenerate solution, we enforce that at least one non-degenerate non-boundary triangle from {each} shape $\mathcal{X}$ and $\mathcal{Y}$ are matched.
These non-degenerate non-boundary triangle-to-triangle matchings are encoded in $\mathbf{\Gamma}_{\text{nb}} \subset \mathbf{\Gamma}$.
To ensure that at least one of these matchings exists, we constrain
\begin{equation}
    \mathbf{1}^T \mathbf{\Gamma}_{\text{nb}} \geq 1.
  \end{equation}

\paragraph{Neighboring Constraint.}
The use of the previous constraint would lead to exactly one matching (match one non-boundary triangle of shape $\mathcal{X}$ to one non-boundary triangle of shape $\mathcal{Y}$).
Therefore, we define a constraint that ensures that neighboring non-boundary triangles remain neighbors after matching -- this guarantees \emph{geometric consistency in the interior}.
We define $\mathbf{\mathring{E}}=
(\mathbf{E}^{\mathcal{X}}_{nb} \times
\mathbf{E}^{\mathcal{Y}}_{nb}) \cup (\mathbf{E}^{\mathcal{X}}_{nb} \times
\mathbf{V}^{\mathcal{Y}}_{nb}) \cup (\mathbf{V}^{\mathcal{X}}_{nb} \times
\mathbf{E}^{\mathcal{Y}}_{nb}) $ as the set of interior product edges.
For a given product triangle $\mathbf{f}_i\in\mathbf{F}$,
we assign the orientation $\{-1,1\}$ to interior product edges that are adjacent to product triangles $\mathbf{f}_j\in\mathbf{F}$, and orientation $0$ if they do not coincide.
This can be achieved by applying the product orientation operator $\mathcal{O}$ to edges $\mathbf{e}_i$ and triangles $\mathbf{f}_j$ in the $4$-dimensional
product space (cf.~\cite{windheuser2011geometrically} for further insights).
We define the \emph{product orientation matrix}
$\bm{\partial}\in\{-1,0,1\}^{|\mathbf{\mathring{E}}|\times|\mathbf{F}|}$ as
\begin{equation}
\label{eq:oriention-prod}
\partial_{i,j}:=\mathcal{O}(\mathbf{e}_i,\mathbf{f}_j),
\end{equation}
By setting $\bm{\partial} \mathbf{\Gamma} = \boldsymbol{0}$, we can ensure that the neighboring constraint is fulfilled, and, consequently, non-boundary triangles are matched in a geometrically consistent fashion.

\paragraph{Cost Function.}
Most feature extractors return vertex-wise features, i.e., a feature matrix $\mathbf{W}^{(\mathcal{X})}\subset\mathbb{R}^{|\mathbf{V}^{(\mathcal{X})}| \times d }$ which contains a $d$-dimensional feature vector for every vertex in $\mathbf{V}^{(\mathcal{X})}$.
Since our algorithm operates on triangles, we set up a matrix $\mathbf{U}^{(\mathcal{X})}\subset\mathbb{R}^{|\mathbf{\bar{F}}^{(\mathcal{X})}| \times d }$ that contains a feature vector for every triangle in $\mathbf{\bar{F}}^{(\mathcal{X})}$.
For every triangle, the features are computed as the {mean of the features of its vertices}.
We define the cost vector {$\mathbf{C} \in\mathbb{R}^{|\mathbf{F}|}$,}
which contains the matching costs for each product triangle. The cost $c_i$ to match a product triangle $\mathbf{f}_i=(\mathbf{f}_k^{(\mathcal{X})},\mathbf{f}_l^{(\mathcal{Y})})\in\mathbf{F}$ is defined by the L2 norm of the features $\mathbf{u}_k^{(\mathcal{X})}, \mathbf{u}_l^{(\mathcal{Y})}$, i.e.~$c_i = ||\mathbf{u}_k^{(\mathcal{X})} - \mathbf{u}_l^{(\mathcal{Y})}||_2$.
The cost vector $\mathbf{C}$ is normalized such that all $c_i \in [0,1]$.
To ensure that the algorithm
{is not biased towards a small number of matchings,}
we further normalize the objective function by the number of elements that are matched in total, i.e., we minimize the mean cost instead of the sum.
Our integer program reads
    \begin{align}
        \label{eq:objective}
        ~&~\underset{\mathbf{\Gamma} \in \{0, 1\}^{|F|}}{\min} \frac{\bigl\langle\mathbf{C},\mathbf{\Gamma}\bigr\rangle}{\mathbf{1}^T \mathbf{\Gamma}} \\
        \text{ s.t. } ~&~ \bm{\partial} \mathbf{\Gamma} = \mathbf{0}, \bm{\pi}^{(\mathcal{X})} \mathbf{\Gamma} \leq \mathbf{1}, \bm{\pi}^{(\mathcal{Y})}\mathbf{\Gamma} \leq \mathbf{1}, \mathbf{1}^T \mathbf{\Gamma}_{\text{nb}}  \geq 1. \nonumber
    \end{align}
A further difficulty, in addition to the binary constraints, is that 
the division by the sum over all entries of $\mathbf{\Gamma}$ makes the objective non-linear.
We solve this integer fractional  minimization problem by means of a sequence of corresponding integer linear programs.
To be specific, we formulate every integer linear program by adding a constraint that fixes the number of elements that have value $1$ in $\mathbf{\Gamma}$ to a number $x \in \{1, 2, 3, \ldots, (|\mathbf{F}^{(\mathcal{X})}| +|\mathbf{F}^{(\mathcal{Y})}|)\}$.
As such, {for fixed $x$}, the integer linear program reads
    \begin{align}
    \label{eq:ilp}
        ~&~\underset{\mathbf{\Gamma} \in \{0, 1\}^{|F|}}{\min} \bigl\langle\mathbf{C},\mathbf{\Gamma}\bigr\rangle \\
        \text{ s.t. } ~&~ \bm{\partial} \mathbf{\Gamma} = \mathbf{0}, \bm{\pi}^{(\mathcal{X})} \mathbf{\Gamma} \leq \mathbf{1}, \bm {\pi}^{(\mathcal{Y})}\mathbf{\Gamma} \leq \mathbf{1},\mathbf{1}^T \mathbf{\Gamma}_{\text{nb}} \geq 1, \nonumber\\
        ~&~ \mathbf{1}^T \mathbf{\Gamma} = x.
\nonumber
    \end{align}
By iterating over all $x$ and comparing the objectives (normalized by $x$), we can find the globally optimal solution of the fractional minimization problem \eqref{eq:objective}.

\subsection{{Search Space Reduction}}
\begin{figure}
\includegraphics[width=1\linewidth]{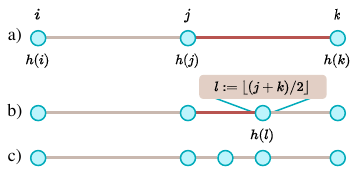}
\caption{Illustration of our \textbf{pruning algorithm}: a) We first calculate an ILP for each initial value for $x$.
Then, we identify intervals with potentially smaller $h(x)$ than the current minimum.
 b) We define new points in search space for $x$ and calculate
 ILPs for these points only. c) We continue till all necessary intervals are solved.}
\label{fig:intervalSearch}
\end{figure}
We define {the minimum objective value of the optimization problem \eqref{eq:ilp} for a specific $x$ as $g(x)$.}
The normalized objective value is $h(x) := \frac{g(x)}{x}$.

To find $x^*$ that minimizes $h(x)$, we can, in principle, compute $g(x)$ for all possible $x \in \mathbf{I} = \{1, \ldots, |\mathbf{F}^{(\mathcal{X})}| + |\mathbf{F}^{(\mathcal{Y})}|\}$, and then select $x^*$ according to the smallest value of $h$.
We avoid exhaustively solving all linear programs by employing a pruning algorithm. 
To this end, we choose a set of starting points {$\mathbf{S} \subset \mathbf{I}$
and solve the optimization problem~\eqref{eq:ilp} 
for all $x \in \mathbf{S}$.}
For every interval given by two consecutive points $j,k \in \mathbf{S}$, we check {whether the set $[j,k] \cap \mathbb{N}$ may potentially contain smaller values than}
$\min_{x' \in \mathbf{S}} h\left(x'\right)$.

To identify those intervals with smaller $h(x)$, we analyse the difference between $g\left(x\right)$ and $g\left(x+1\right)$.
Increasing $x$ by one is equivalent to allowing an additional product triangle in the matching.
This product triangle's cost is at most one (since $c_i \leq 1$ for $c_i \in \mathbf{C}$).
Moreover, this additional triangle can always be matched to a boundary edge, as boundary edges have no orientation constraints. 
Thus the maximal cost of adding one product triangle is bounded by $1$.

Hence, we know $g(x+1) \leq g(x) + 1$, which gives rise to $g(x+1) - g(x) \leq 1$. 
We observe the similarity between our bound and the Lipschitz constant of continuous functions.
It follows that $g\left(k\right) - g\left(j\right) \leq k-j$ for any set $[j,k] \cap \mathbb{N}$
,  and $h\left(k\right) - h\left(j\right) \leq \sum_{i=j}^{k-1} \frac{1}{i}$ (see supplementary).
Finally, for a set $[j,k] \cap \mathbb{N}$ the potentially minimal objective is given as
$h(j)-h(k)-\sum_{i=j}^{k-1} \frac{1}{i}$.
If this value is smaller than $\min_{x' \in \mathbf{I}} h\left(x'\right)$, we add $l :=  \lfloor \frac{j+k}{2} \rfloor$ to $\mathbf{S}$.
We repeat this process until all intervals are determined to be impossible to contain smaller minima or all possible values for $x$ have been visited, i.e., $\mathbf{S} = \{1, \ldots, |\mathbf{F}^{(\mathcal{X})}| + |\mathbf{F}^{(\mathcal{Y})}|\}$.
See Figure~\ref{fig:intervalSearch} for an illustration of our pruning algorithm.

\subsection{Implementation Details}
\subsubsection{Upsampling to Original Resolution}
\begin{figure}
\includegraphics[width=1\linewidth]{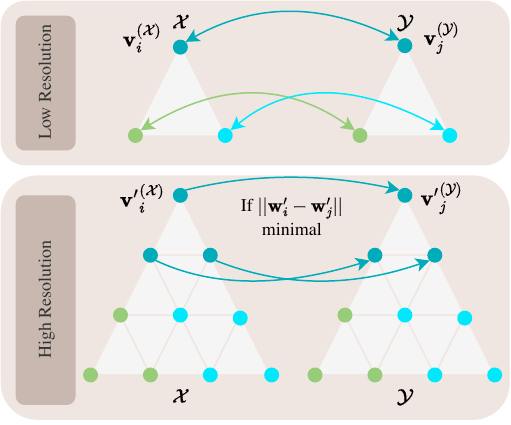}
\caption{Our \textbf{upsampling strategy} via nearest neighbor search: Given a matching between vertices $\mathbf{v}^{(\mathcal{X})}_i$ and $\mathbf{v}^{(\mathcal{Y})}_j$ on lower resolution, we determine the matchings in high resolution by only considering
the upsampled vertices (dark blue).
A vertex $\mathbf{v'}^{(\mathcal{X})}_i$, which was downsampled on $\mathbf{v}^{(\mathcal{X})}_i$ can only be matched on the
upsampled vertices of $\mathbf{v}^{(\mathcal{Y})}_j$ (dark blue vertices in $\mathcal{Y}$).
We choose this vertex $\mathbf{v'}^{(\mathcal{Y})}_j$
that has the smallest distance in feature space of $\mathbf{W}^{(\mathcal{X})}$ and $\mathbf{W}^{(\mathcal{Y})}$.
}
\label{fig:low_to_high}
\end{figure}
{To achieve reasonable running times, we perform a coarse-to-fine processing, in which we solve Problem~\eqref{eq:objective} for meshes with lower resolution and then upsample the low-resolution matching to a high-resolutions matching.}
Let $\mathbf{v}^{(\mathcal{X})}_i \leftrightarrow \mathbf{v}^{(\mathcal{Y})}_j$ be the vertex matchings of the low resolution shapes.
{For upsampling, we ensure
that upsampled vertices of $\mathbf{v}^{(\mathcal{X})}_i$ in shape $\mathcal{X}$ are matched to upsampled vertices of the corresponding low resolution vertex $\mathbf{v}^{(\mathcal{Y})}_j$ in shape $\mathcal{Y}$.}
We perform a nearest neighbor search of the given vertex features $\mathbf{W}^{(\mathcal{X})}$ and $\mathbf{W}^{(\mathcal{Y})}$ in this reduced search space.
We show an illustration in Figure~\ref{fig:low_to_high}.

\subsubsection{Runtime versus Optimality Trade-Off}
\label{sec:heuristic}
While our algorithm leads to globally optimal solutions, it has exponential worst-case runtime. 
To reduce the runtime, we set a time limit to solve the individual ILP subproblems.

In \cite{thunberg2023non}, convergence speed is used as criterion for model selection -- inspired by this, we propose a similar heuristic that is based on our observation that finding the solution that leads to the minimum mean is in many cases the fastest to compute among all values of $x$ in problem \eqref{eq:ilp}.
Thus, we can reduce the time budget for the ILP subproblems and select the solution with smallest objective among all subproblems finished within the time limit.
If no subproblem terminates within the time budget, we increase the time budget gradually.

\section{Experimental Results}
In the following, we experimentally evaluate our proposed method.
We show that our method outperforms current SOTA partial-to-partial shape matching methods in terms of Intersection over Union~(IoU)
and Geodesic Error. Further ablation studies can be found in the supplementary.

\subsection{Datasets}
In our experiments, we use two different datasets:
\paragraph{Intra-Class Dataset}
First, we use a modified version of the CUTS Partial-to-Partial dataset (\textsc{CP2P})~\cite{attaiki2021dpfm}.
The dataset is defined on the partial shapes of the SHREC16 \textsc{CUTS} dataset~\cite{cosmo2016shrec}, which include animals and human shapes in different poses.
In \cite{attaiki2021dpfm} the authors introduce the \textsc{CP2P} dataset by splitting the original training set into their training and test set.
Instead, we use samples from the original SHREC16 CUTS test set to generate our test set \textsc{CP2P TEST'24}, such that we can utilize the whole SHREC16 train set for training.
During this work we have observed that the original SHREC16 CUTS test split contains 47 shapes that are almost identical (up to some slight vertex jitter) to shapes in the original training set (see supplementary material for details). 
Thus, we consider only the remaining 153 shapes of the original test set.
We propose to use this subset of shapes as test set in future research. We make this new test split publicly available.
Similar to the version provided in~\cite{attaiki2021dpfm}, the randomly chosen samples have an overlapping region between 10-90\%.
As in~\cite{attaiki2021dpfm}, we choose 50 random sample {pairs} and evaluate them in both directions to obtain 100 matching {pairs.}

\paragraph{Inter-Class Dataset}
While the \textsc{CP2P} dataset provides intra-class matching, we present a new dataset called PARTIALSMAL that allows the evaluation of non-isometric inter-class shape matching.
The dataset is based on the SMAL dataset~\cite{Zuffi:CVPR:2017}, which includes 49 animals from 8 species.
Similar to~\cite{cao2023unsupervised}, we use a train/test split in different species, such that the test set contains three species that are not included in the train set that contains five species.
Similar to~\cite{donati2022DeepCFMaps}, we re-mesh our dataset as presented in~\cite{ren2018continuous}, such that methods cannot easily overfit the connectivity of the meshes. We cut the shapes in parts by configuring planes with normal vector $(x,y,z)$, with $(x,y,z) \in \{-1,0,1\}^3 \backslash (0,0,0)$.
This results in 26 different configurations for every shape.
See supplementary for more information. 

\subsection{Experimental Setup}
We compute the features required to compute matching costs (for our method and Sm-Comb~\cite{roetzer2022scalable})
with~\cite{cao2023unsupervised} based on the full-to-full matching training set of SMAL~\cite{Zuffi:CVPR:2017} and the partial-to-full training set of SHREC16 CUTS~\cite{cosmo2016shrec}, respectively.
During this work, the authors of~\cite{bracha2023partial} pointed out that for SHREC16 the test and training set used in~\cite{cao2023unsupervised} are not completely independent:
TOSCA is used for pre-training of~\cite{cao2023unsupervised}, while the SHREC16 dataset was constructed from TOSCA by first remeshing and then introducing missing parts. To avoid data leakage, we instead pretrain~\cite{cao2023unsupervised} using a combination of full shapes from the DT4D~\cite{magnet2022smooth, li20214dcomplete}, SMAL~\cite{Zuffi:CVPR:2017}, FAUST~\cite{bogo2014faust} and SCAPE~\cite{anguelov2005scape} datasets\footnote{Cao~\etal now provide additional results with similar pretraining, substantiating their original claims in~\cite{cao2023unsupervised}.}.
During test time, no full shapes are needed. Also, notice that full shapes of the same species of the SMAL test set have not been seen during training as the train and test sets contain different animal species.
We use Gurobi~\cite{gurobi} interfaced through YALMIP~\cite{Lofberg2004} to solve the linear integer programs.
We set the maximum time limit for every shape pair to 10 hours. 
Unless stated otherwise, the time limit for one ILP is set to 15 minutes.
The initial interval size for the pruning algorithm is 50. 
If no solution is found, we first reduce the initial interval size to 25. 
If still no matching is found, we further downsample the meshes.
For more information on how to prepare the shapes, see the supplementary material.

\subsection{Comparison Methods}
We compare our work with the two existing partial-to-partial shape matching approaches: Sm-comb~\cite{roetzer2022scalable} and DPFM~\cite{attaiki2021dpfm}.
For Sm-comb~\cite{roetzer2022scalable}, we close every hole by adding one vertex in the middle of a hole and then connect this vertex with the vertices on the respective boundary, such that we obtain two full water-tight shapes, as done in the original paper.
The mean of the feature vectors of all vertices along this boundary defines the feature of the added vertex.
To allow for a fair comparison, we
use the same SOTA features to compute matching costs for our method and Sm-comb.
The supervised method DPFM~\cite{attaiki2021dpfm} is readily applicable to \textsc{CP2P TEST}.
For PARTIALSMAL, we provide a train set that is similarly generated as the test set.
We cut the full shapes of the SMAL~\cite{Zuffi:CVPR:2017} and provide a total of 304 random combinations of these for training DPFM (same amount as in the CP2P train set).

\newcommand{\includeIoU}[1]{ \includegraphics[width=0.18\linewidth]{vis/cuts_cat_shape_13_cuts_cat_shape_7/#1.png}
}
\newcommand{\includeIoUHorseSmal}[1]{ \includegraphics[width=0.12\linewidth]{vis/ious_qualitative_results/#1.png}
}
\newcommand{\includeIoUSrcWolf}[1]{ \includegraphics[width=0.12\linewidth]{vis/cuts_cat_shape_13_cuts_cat_shape_7/#1.png}
}

\newcommand{\includeIoUSmalCougar}[1]{ \includegraphics[width=0.08\linewidth]{vis/ious_qualitative_results/#1.png}
}

\newcommand{\includeIoUHorse}[1]{ \includegraphics[width=0.18\linewidth]{vis/cuts_horse_shape_11_cuts_horse_shape_5/#1.png}
}
\newcommand{\includeIoUSmalHorseHippo}[1]{ \includegraphics[width=0.08\linewidth]{vis/cuts_11_horse_08_cuts_19_hippo_03/#1.png}
}

\newcommand{\includeIoUSmalHorseHippoSrc}[1]{ \includegraphics[width=0.05\linewidth]{vis/cuts_11_horse_08_cuts_19_hippo_03/#1.png}
}

\newcommand{\includeIoUHorseSrc}[1]{ \includegraphics[width=0.12\linewidth]{vis/cuts_horse_shape_11_cuts_horse_shape_5/#1.png}
}

\newcommand{\includeIoURow}[1]{  \includeIoUSrcWolf{shrec16_direct_False_new_gt_y_#1_overlap}
& \includeIoU{shrec16_direct_False_new_gt_x_#1_overlap} &
\includeIoU{shrec16_double_closed_direct_False_new_#1_overlap} &
\includeIoU{results_dpfm_new_#1_overlap}
&
\includeIoU{shrec16_direct_False_new_#1_overlap}
}

\newcommand{\includeIoUHorseRow}[1]{  \includeIoUHorseSrc{shrec16_direct_False_new_gt_y_#1_overlap}
& \includeIoUHorse{shrec16_direct_False_new_gt_x_#1_overlap} &
\includeIoUHorse{shrec16_double_closed_direct_False_new_#1_overlap} &
\includeIoUHorse{results_dpfm_new_#1_overlap}
&
\includeIoUHorse{shrec16_direct_False_new_#1_overlap}
}

\newcommand{\includeIoURowSmal}[1]{  \includeIoUHorseSmal{partialSmal_direct_False_gt_y_#1_overlap}
& \includeIoUSmalCougar{partialSmal_direct_False_gt_x_#1_overlap} &
\includeIoUSmalCougar{smal_double_closed_direct_False_#1_overlap} &
\includeIoUSmalCougar{smal_dpfm_#1_overlap}
&
\includeIoUSmalCougar{partialSmal_direct_False_#1_overlap}
}
\newcommand{\includeIoURowSmalHorseHippo}[1]{  \includeIoUSmalHorseHippoSrc{partialSmal_direct_False_gt_y_#1_overlap}
& \includeIoUSmalHorseHippo{partialSmal_direct_False_gt_x_#1_overlap} &
\includeIoUSmalHorseHippo{smal_double_closed_direct_False_#1_overlap} &
\includeIoUSmalHorseHippo{smal_dpfm_#1_overlap}
&
\includeIoUSmalHorseHippo{partialSmal_direct_False_#1_overlap}
}
\begin{figure*}[htpb]
    \centering
    \setlength\tabcolsep{+2pt} %
        \begin{tabular}{c | c c c c c}
            \centering
             &Source & Ground Truth &  Sm-comb~\cite{roetzer2022scalable} & DPFM~\cite{attaiki2021dpfm} & \textbf{Ours} \\
             \hline
             \multirow{2}{*}[1.2em]{\rotatebox{90}{\textsc{CP2P TEST}}}&
\includeIoURow{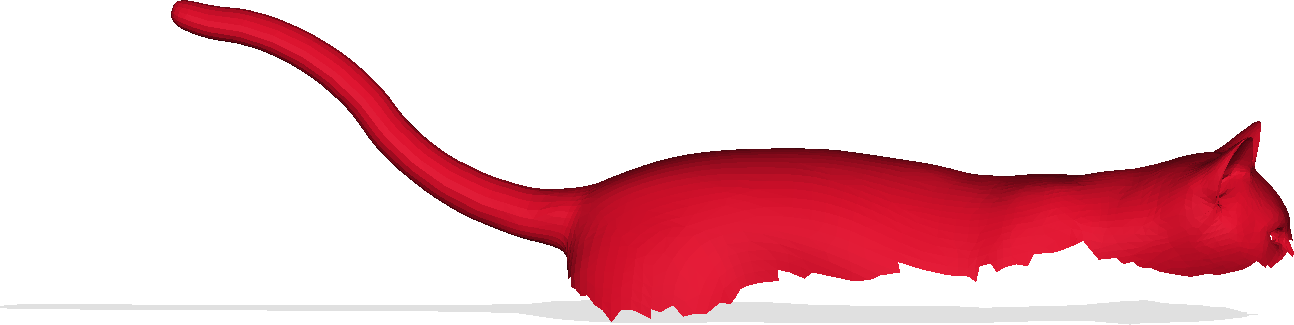}
\\
&\includeIoUHorseRow{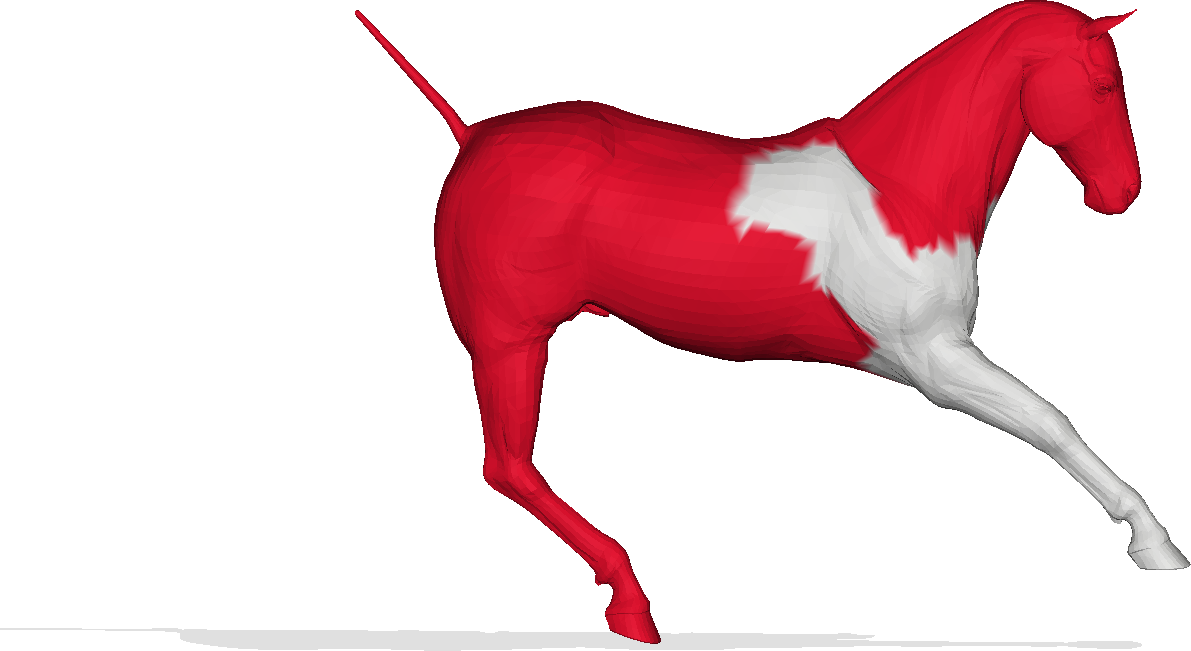}
\\
\hline
\multirow{2}{*}[1.2em]{\rotatebox{90}{PARTIALSMAL}}&\includeIoURowSmal{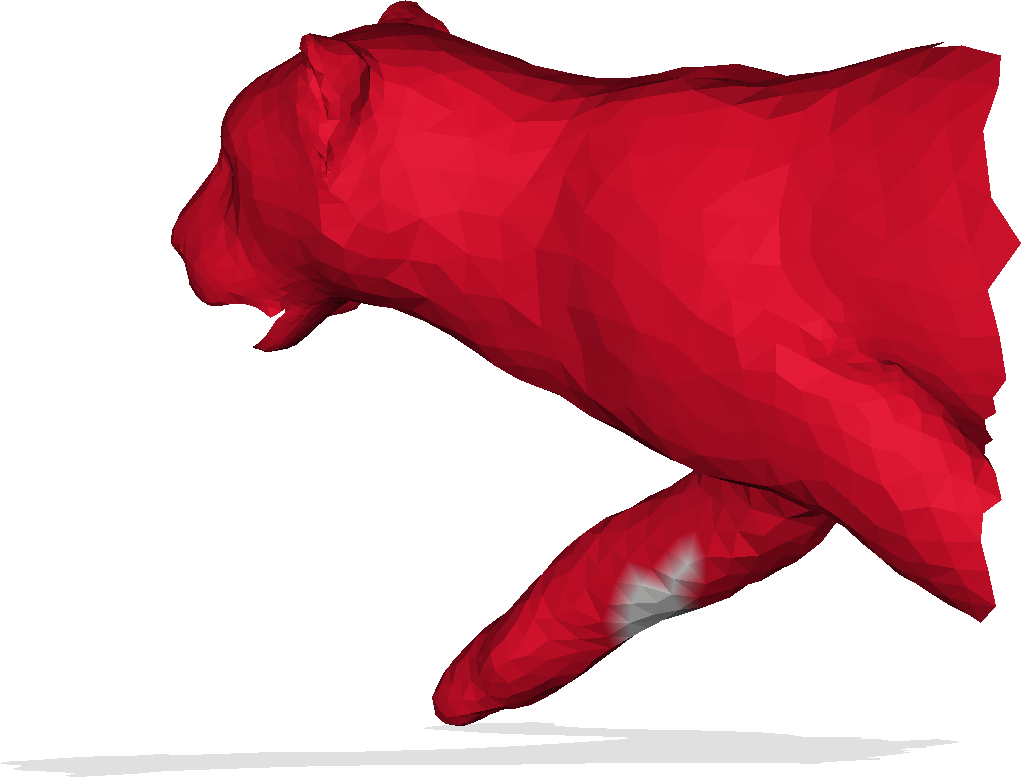}
\\
&\includeIoURowSmalHorseHippo{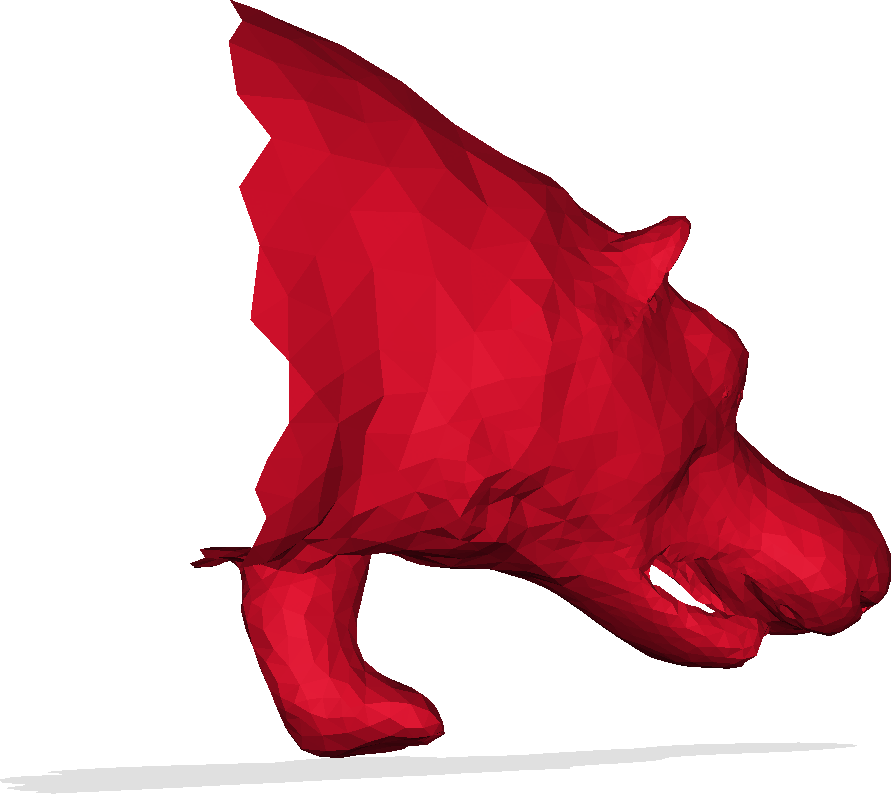}
\\
        \end{tabular}
    \caption{Comparisons of \textbf{overlapping region predictions} on CP2P TEST and PARTIALSMAL.
    DPFM~\cite{attaiki2021dpfm} often returns noisy predictions and Sm-comb~\cite{roetzer2022scalable} predicts that the shape is matched completely or at the wrong position in the given example shapes (first two rows CP2P TEST, last two rows PARTIALSMAL).
    Only our method returns results close to the expected ground truth. We denote that our method performs overlapping prediction and matching simultaneously.
    \label{fig:iou_qualitative}
    }
\end{figure*}

\subsection{Overlapping Region Prediction}
\begin{figure}[htpb]%
    \setlength\tabcolsep{-5pt} 
    \begin{tabular}{ll}
      \newcommand{\pckLineWidth}{3pt}
\newcommand{\plotWidth}{1*\linewidth}
\newcommand{\plotHeight}{0.75*\linewidth}
\newcommand{\pckTitle}{\textsc{CP2P TEST}}

\pgfplotsset{%
    label style = {font=\LARGE},
    tick label style = {font=\large},
    title style =  {font=\LARGE},
    legend style={  fill= gray!10,
                    fill opacity=0.6, 
                    font=\LARGE,
                    draw=gray!20, 
                    text opacity=1}
}
\begin{tikzpicture}[scale=0.45, transform shape]
	\begin{axis}[
		width=\plotWidth,
		height=\plotHeight,
		grid=major,
		title=\pckTitle,
		legend style={
			at={(0.03,0.03)},
			anchor=south west,
			legend columns=1},
		legend cell align={left},
	ylabel={{\LARGE
        $\%$ Samples $>$ IoU}},
        xmin=0,
        xmax=100,
        xlabel=\LARGE
        IoU (x100),
        ylabel near ticks,
        xtick={0, 20, 40, 60, 80, 100},
        xticklabels={$0$, $20$, $40$, $60$, $80$, $100$},
        ymin=0,
        ymax=100,
        ytick={0, 20, 40, 60, 80, 100},
        yticklabels={$0$, $20$, $40$, $60$, $80$, $100$},
	]
 \addplot [color=cPLOTroetzer, dashed, smooth, line width=\pckLineWidth]
    table[row sep=crcr]{%
0  100.0\\
1  96.46017699115043\\
2  96.46017699115043\\
3  96.46017699115043\\
4  94.69026548672566\\
5  94.69026548672566\\
6  92.92035398230088\\
7  92.92035398230088\\
8  89.38053097345133\\
9  89.38053097345133\\
10  88.49557522123894\\
11  87.61061946902655\\
12  87.61061946902655\\
13  87.61061946902655\\
14  87.61061946902655\\
15  84.95575221238938\\
16  83.1858407079646\\
17  83.1858407079646\\
18  82.30088495575221\\
19  82.30088495575221\\
20  82.30088495575221\\
21  82.30088495575221\\
22  81.41592920353983\\
23  81.41592920353983\\
24  80.53097345132744\\
25  78.76106194690266\\
26  77.87610619469027\\
27  76.99115044247787\\
28  75.22123893805309\\
29  71.68141592920354\\
30  71.68141592920354\\
31  71.68141592920354\\
32  69.02654867256636\\
33  68.14159292035397\\
34  67.2566371681416\\
35  66.3716814159292\\
36  65.48672566371681\\
37  61.94690265486725\\
38  61.06194690265486\\
39  58.4070796460177\\
40  56.63716814159292\\
41  55.75221238938053\\
42  54.86725663716814\\
43  52.21238938053098\\
44  52.21238938053098\\
45  50.442477876106196\\
46  49.557522123893804\\
47  46.902654867256636\\
48  46.017699115044245\\
49  44.24778761061947\\
50  43.36283185840708\\
51  42.47787610619469\\
52  41.5929203539823\\
53  40.707964601769916\\
54  38.93805309734513\\
55  38.93805309734513\\
56  37.16814159292036\\
57  37.16814159292036\\
58  35.39823008849557\\
59  33.6283185840708\\
60  31.858407079646017\\
61  30.973451327433626\\
62  30.08849557522124\\
63  29.20353982300885\\
64  29.20353982300885\\
65  27.43362831858407\\
66  26.548672566371685\\
67  25.663716814159294\\
68  25.663716814159294\\
69  25.663716814159294\\
70  23.893805309734514\\
71  22.123893805309734\\
72  22.123893805309734\\
73  22.123893805309734\\
74  20.353982300884958\\
75  15.929203539823009\\
76  15.04424778761062\\
77  14.15929203539823\\
78  14.15929203539823\\
79  12.389380530973451\\
80  11.504424778761061\\
81  9.734513274336283\\
82  9.734513274336283\\
83  9.734513274336283\\
84  9.734513274336283\\
85  8.849557522123893\\
86  7.964601769911504\\
87  5.3097345132743365\\
88  5.3097345132743365\\
89  5.3097345132743365\\
90  5.3097345132743365\\
91  5.3097345132743365\\
92  5.3097345132743365\\
93  4.424778761061947\\
94  3.5398230088495577\\
95  2.6548672566371683\\
96  0.8849557522123894\\
97  0.8849557522123894\\
98  0.0\\
99  0.0\\
100  0.0\\
    };
    \addlegendentry{\textcolor{black}{Sm-comb}}
    \addplot [color=cPLOTdpfm, dashed, smooth, line width=\pckLineWidth]
    table[row sep=crcr]{%
0  100.0\\
1  100.0\\
2  100.0\\
3  99.0\\
4  99.0\\
5  99.0\\
6  99.0\\
7  98.0\\
8  98.0\\
9  98.0\\
10  98.0\\
11  97.0\\
12  97.0\\
13  97.0\\
14  96.0\\
15  96.0\\
16  96.0\\
17  96.0\\
18  95.0\\
19  94.0\\
20  93.0\\
21  92.0\\
22  92.0\\
23  91.0\\
24  90.0\\
25  89.0\\
26  89.0\\
27  89.0\\
28  89.0\\
29  89.0\\
30  89.0\\
31  89.0\\
32  87.0\\
33  83.0\\
34  82.0\\
35  82.0\\
36  79.0\\
37  77.0\\
38  76.0\\
39  73.0\\
40  72.0\\
41  71.0\\
42  68.0\\
43  68.0\\
44  63.0\\
45  62.0\\
46  59.0\\
47  59.0\\
48  57.99999999999999\\
49  57.99999999999999\\
50  57.99999999999999\\
51  56.99999999999999\\
52  54.0\\
53  54.0\\
54  53.0\\
55  53.0\\
56  51.0\\
57  50.0\\
58  48.0\\
59  46.0\\
60  44.0\\
61  42.0\\
62  39.0\\
63  39.0\\
64  37.0\\
65  37.0\\
66  32.0\\
67  32.0\\
68  31.0\\
69  30.0\\
70  28.000000000000004\\
71  26.0\\
72  25.0\\
73  25.0\\
74  24.0\\
75  24.0\\
76  23.0\\
77  21.0\\
78  19.0\\
79  17.0\\
80  15.0\\
81  14.000000000000002\\
82  13.0\\
83  13.0\\
84  11.0\\
85  9.0\\
86  6.0\\
87  6.0\\
88  6.0\\
89  6.0\\
90  4.0\\
91  2.0\\
92  2.0\\
93  2.0\\
94  1.0\\
95  1.0\\
96  0.0\\
97  0.0\\
98  0.0\\
99  0.0\\
100  0.0\\
    };
    \addlegendentry{\textcolor{black}{DPFM}}
    \addplot [color=cPLOTours, smooth, line width=\pckLineWidth]
    table[row sep=crcr]{%
0  100.0\\
1  100.0\\
2  100.0\\
3  100.0\\
4  100.0\\
5  100.0\\
6  100.0\\
7  100.0\\
8  100.0\\
9  100.0\\
10  100.0\\
11  100.0\\
12  100.0\\
13  100.0\\
14  100.0\\
15  100.0\\
16  99.0\\
17  98.0\\
18  98.0\\
19  98.0\\
20  97.0\\
21  97.0\\
22  96.0\\
23  93.0\\
24  93.0\\
25  93.0\\
26  93.0\\
27  93.0\\
28  93.0\\
29  93.0\\
30  91.0\\
31  90.0\\
32  89.0\\
33  89.0\\
34  89.0\\
35  86.0\\
36  86.0\\
37  85.0\\
38  84.0\\
39  83.0\\
40  81.0\\
41  81.0\\
42  81.0\\
43  80.0\\
44  80.0\\
45  79.0\\
46  76.0\\
47  76.0\\
48  75.0\\
49  74.0\\
50  73.0\\
51  72.0\\
52  72.0\\
53  72.0\\
54  72.0\\
55  72.0\\
56  71.0\\
57  70.0\\
58  70.0\\
59  68.0\\
60  67.0\\
61  65.0\\
62  65.0\\
63  65.0\\
64  64.0\\
65  64.0\\
66  64.0\\
67  64.0\\
68  63.0\\
69  62.0\\
70  62.0\\
71  59.0\\
72  56.99999999999999\\
73  54.0\\
74  54.0\\
75  52.0\\
76  50.0\\
77  48.0\\
78  45.0\\
79  44.0\\
80  41.0\\
81  41.0\\
82  41.0\\
83  39.0\\
84  38.0\\
85  35.0\\
86  34.0\\
87  33.0\\
88  28.999999999999996\\
89  27.0\\
90  25.0\\
91  23.0\\
92  23.0\\
93  22.0\\
94  20.0\\
95  17.0\\
96  10.0\\
97  8.0\\
98  5.0\\
99  0.0\\
100  0.0\\
    };
    \addlegendentry{\textcolor{black}{Ours}}
	\end{axis}
\end{tikzpicture}   & \newcommand{\pckLineWidth}{3pt}
\newcommand{\plotWidth}{1*\linewidth}
\newcommand{\plotHeight}{0.75*\linewidth}
\newcommand{\pckTitle}{\textsc{PARTIALSMAL}}

\pgfplotsset{%
    label style = {font=\LARGE},
    tick label style = {font=\large},
    title style =  {font=\LARGE},
    legend style={  fill= gray!10,
                    fill opacity=0.6, 
                    font=\LARGE,
                    draw=gray!20, 
                    text opacity=1}
}
\begin{tikzpicture}[scale=0.45, transform shape]
	\begin{axis}[
		width=\plotWidth,
		height=\plotHeight,
		grid=major,
		title=\pckTitle,
		legend style={
			at={(0.03,0.03)},
			anchor=south west,
			legend columns=1},
		legend cell align={left},
	ylabel={{\LARGE
        $\%$ Samples $>$ IoU}},
        xmin=0,
        xmax=100,
        xlabel=\LARGE
        IoU (x100),
        ylabel near ticks,
        xtick={0, 20, 40, 60, 80, 100},
        xticklabels={$0$, $20$, $40$, $60$, $80$, $100$},
        ymin=0,
        ymax=100,
        ytick={0, 20, 40, 60, 80, 100},
        yticklabels={$0$, $20$, $40$, $60$, $80$, $100$},
	]
 \addplot [color=cPLOTroetzer, dashed, smooth, line width=\pckLineWidth]
    table[row sep=crcr]{%
0  100.0\\
1  99.0\\
2  98.0\\
3  98.0\\
4  98.0\\
5  98.0\\
6  98.0\\
7  96.0\\
8  96.0\\
9  96.0\\
10  96.0\\
11  96.0\\
12  95.0\\
13  95.0\\
14  95.0\\
15  94.0\\
16  94.0\\
17  93.0\\
18  93.0\\
19  93.0\\
20  92.0\\
21  92.0\\
22  91.0\\
23  91.0\\
24  89.0\\
25  89.0\\
26  87.0\\
27  86.0\\
28  86.0\\
29  84.0\\
30  83.0\\
31  80.0\\
32  77.0\\
33  77.0\\
34  75.0\\
35  75.0\\
36  73.0\\
37  73.0\\
38  72.0\\
39  71.0\\
40  69.0\\
41  67.0\\
42  65.0\\
43  65.0\\
44  63.0\\
45  61.0\\
46  61.0\\
47  60.0\\
48  57.99999999999999\\
49  57.99999999999999\\
50  56.99999999999999\\
51  55.00000000000001\\
52  53.0\\
53  52.0\\
54  50.0\\
55  49.0\\
56  47.0\\
57  47.0\\
58  47.0\\
59  46.0\\
60  46.0\\
61  45.0\\
62  42.0\\
63  40.0\\
64  37.0\\
65  36.0\\
66  36.0\\
67  35.0\\
68  34.0\\
69  32.0\\
70  32.0\\
71  30.0\\
72  30.0\\
73  30.0\\
74  28.999999999999996\\
75  28.000000000000004\\
76  28.000000000000004\\
77  25.0\\
78  24.0\\
79  22.0\\
80  22.0\\
81  20.0\\
82  20.0\\
83  18.0\\
84  18.0\\
85  16.0\\
86  15.0\\
87  15.0\\
88  10.0\\
89  9.0\\
90  5.0\\
91  4.0\\
92  3.0\\
93  2.0\\
94  1.0\\
95  1.0\\
96  0.0\\
97  0.0\\
98  0.0\\
99  0.0\\
100  0.0\\
    };
    \addlegendentry{\textcolor{black}{Sm-comb}}
    \addplot [color=cPLOTdpfm, dashed, smooth, line width=\pckLineWidth]
    table[row sep=crcr]{%
0  100.0\\
1  100.0\\
2  100.0\\
3  100.0\\
4  100.0\\
5  100.0\\
6  99.0\\
7  99.0\\
8  99.0\\
9  99.0\\
10  99.0\\
11  99.0\\
12  99.0\\
13  98.0\\
14  95.0\\
15  93.0\\
16  92.0\\
17  92.0\\
18  90.0\\
19  90.0\\
20  88.0\\
21  87.0\\
22  84.0\\
23  83.0\\
24  83.0\\
25  83.0\\
26  79.0\\
27  77.0\\
28  76.0\\
29  73.0\\
30  71.0\\
31  70.0\\
32  69.0\\
33  69.0\\
34  67.0\\
35  66.0\\
36  64.0\\
37  62.0\\
38  60.0\\
39  60.0\\
40  57.99999999999999\\
41  57.99999999999999\\
42  56.00000000000001\\
43  54.0\\
44  50.0\\
45  46.0\\
46  43.0\\
47  43.0\\
48  41.0\\
49  41.0\\
50  41.0\\
51  39.0\\
52  37.0\\
53  36.0\\
54  35.0\\
55  35.0\\
56  34.0\\
57  34.0\\
58  34.0\\
59  32.0\\
60  31.0\\
61  31.0\\
62  31.0\\
63  28.999999999999996\\
64  28.999999999999996\\
65  28.999999999999996\\
66  28.000000000000004\\
67  27.0\\
68  26.0\\
69  25.0\\
70  22.0\\
71  22.0\\
72  22.0\\
73  20.0\\
74  20.0\\
75  20.0\\
76  19.0\\
77  18.0\\
78  17.0\\
79  16.0\\
80  14.000000000000002\\
81  14.000000000000002\\
82  12.0\\
83  12.0\\
84  11.0\\
85  10.0\\
86  10.0\\
87  10.0\\
88  7.000000000000001\\
89  6.0\\
90  5.0\\
91  5.0\\
92  5.0\\
93  5.0\\
94  3.0\\
95  3.0\\
96  2.0\\
97  2.0\\
98  2.0\\
99  0.0\\
100  0.0\\
    };
    \addlegendentry{\textcolor{black}{DPFM}}
    \addplot [color=cPLOTours, smooth, line width=\pckLineWidth]
    table[row sep=crcr]{%
0  100.0\\
1  98.0\\
2  98.0\\
3  98.0\\
4  98.0\\
5  98.0\\
6  98.0\\
7  98.0\\
8  98.0\\
9  98.0\\
10  98.0\\
11  98.0\\
12  98.0\\
13  98.0\\
14  98.0\\
15  98.0\\
16  97.0\\
17  97.0\\
18  97.0\\
19  97.0\\
20  97.0\\
21  96.0\\
22  95.0\\
23  95.0\\
24  94.0\\
25  93.0\\
26  92.0\\
27  91.0\\
28  91.0\\
29  89.0\\
30  89.0\\
31  88.0\\
32  88.0\\
33  85.0\\
34  84.0\\
35  84.0\\
36  84.0\\
37  84.0\\
38  82.0\\
39  82.0\\
40  82.0\\
41  81.0\\
42  80.0\\
43  78.0\\
44  77.0\\
45  76.0\\
46  76.0\\
47  76.0\\
48  76.0\\
49  76.0\\
50  74.0\\
51  73.0\\
52  72.0\\
53  71.0\\
54  69.0\\
55  67.0\\
56  66.0\\
57  62.0\\
58  61.0\\
59  59.0\\
60  56.99999999999999\\
61  56.99999999999999\\
62  55.00000000000001\\
63  55.00000000000001\\
64  53.0\\
65  53.0\\
66  51.0\\
67  49.0\\
68  49.0\\
69  47.0\\
70  46.0\\
71  44.0\\
72  44.0\\
73  41.0\\
74  41.0\\
75  41.0\\
76  41.0\\
77  36.0\\
78  35.0\\
79  33.0\\
80  33.0\\
81  32.0\\
82  31.0\\
83  30.0\\
84  27.0\\
85  25.0\\
86  24.0\\
87  23.0\\
88  21.0\\
89  20.0\\
90  20.0\\
91  18.0\\
92  16.0\\
93  14.000000000000002\\
94  12.0\\
95  9.0\\
96  6.0\\
97  4.0\\
98  2.0\\
99  2.0\\
100  0.0\\
    };
    \addlegendentry{\textcolor{black}{Ours}}
	\end{axis}
\end{tikzpicture}
    \end{tabular}
\caption{Comparison of \textbf{IoU scores} on \textsc{CP2P TEST} dataset (left) and PARTIALSMAL dataset (right). Our method outperforms current SOTA partial-to-partial methods~\cite{attaiki2021dpfm, roetzer2022scalable}.} 
\label{fig:ious_comparisons_quant}
\end{figure}
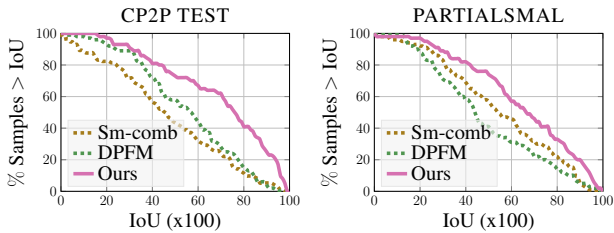

\paragraph{Intersection over Union (IoU)}
Let $P \in \{0,1\}^{(|V| \times 1)}$ and $G \in \{0,1\}^{(|V| \times 1)}$ be   vectors (predicted and ground truth, respectively), that indicate whether the respective vertex is part of the matching.
With that, we quantitatively evaluate the overlapping region based on the 
intersection over union 
$\text{IoU} = \frac{|P \cap GT|}{|P \cup GT|}.$
\paragraph{Results} We compute the IoU for the predicted overlapping region compared to the ground truth. We observe that our method outperforms the two competing methods in terms of the number of shapes that exceed a specific IoU (see Figure~\ref{fig:ious_comparisons_quant}). 
Additionally, we achieve a better mean IoU than the two baselines  (see Table~\ref{tab:objective}). 
In Figure~\ref{fig:iou_qualitative}, we show qualitative results of the overlapping region prediction.

\begin{table}[h]
    \centering
    \begin{tabular}{c||c | c | c}
      Dataset & Sm-comb~\cite{roetzer2022scalable} & DPFM~\cite{attaiki2021dpfm} & Ours\\
      \hline
        \textsc{CP2P TEST} & 
        {57.86}
        & 
        {54.93}
        & 
        {\textbf{69.29}}
        \\
        PARTIALSMAL & 54.76 & 48.31 & \textbf{64.34}
    \end{tabular}
    \caption{Comparison of mean \textbf{IoU} scores (x100). Our method outperforms Sm-comb~\cite{roetzer2022scalable} and DPFM~\cite{attaiki2021dpfm} on both datasets. }
    \label{tab:objective}
\end{table}

\newcommand{\includeGeoErrorCat}[3]{ \includegraphics[width=#2\linewidth]{#3/#1.png}
}

\newcommand{\includeGeoErrorRowCat}[5]{
\includeGeoErrorCat{shrec16_double_closed_direct_False_new_#1_#2_overlap}{#3}{#5}
&
\includeGeoErrorCat{shrec16_double_closed_direct_False_new_#2_#1_overlap}{#4}{#5}
&
\includeGeoErrorCat{results_dpfm_new_#1_#2_overlap}{#3}{#5}
&
\includeGeoErrorCat{results_dpfm_new_#2_#1_overlap}{#4}{#5}&\includeGeoErrorCat{shrec16_direct_False_new_#1_#2_overlap}{#3}{#5}
&
\includeGeoErrorCat{shrec16_direct_False_new_#2_#1_overlap}{#4}{#5}
}
\newcommand{\includeGeoErrorRowSmalNew}[5]{
\includeGeoErrorCat{smal_double_closed_direct_False_#1_#2_overlap}{#3}{#5}
&
\includeGeoErrorCat{smal_double_closed_direct_False_#2_#1_overlap}{#4}{#5}
&
\includeGeoErrorCat{smal_dpfm_#1_#2_overlap}{#3}{#5}
&
\includeGeoErrorCat{smal_dpfm_#2_#1_overlap}{#4}{#5}&\includeGeoErrorCat{partialSmal_direct_False_#1_#2_overlap}{#3}{#5}
&
\includeGeoErrorCat{partialSmal_direct_False_#2_#1_overlap}{#4}{#5}
}

\begin{figure*}[htpb]
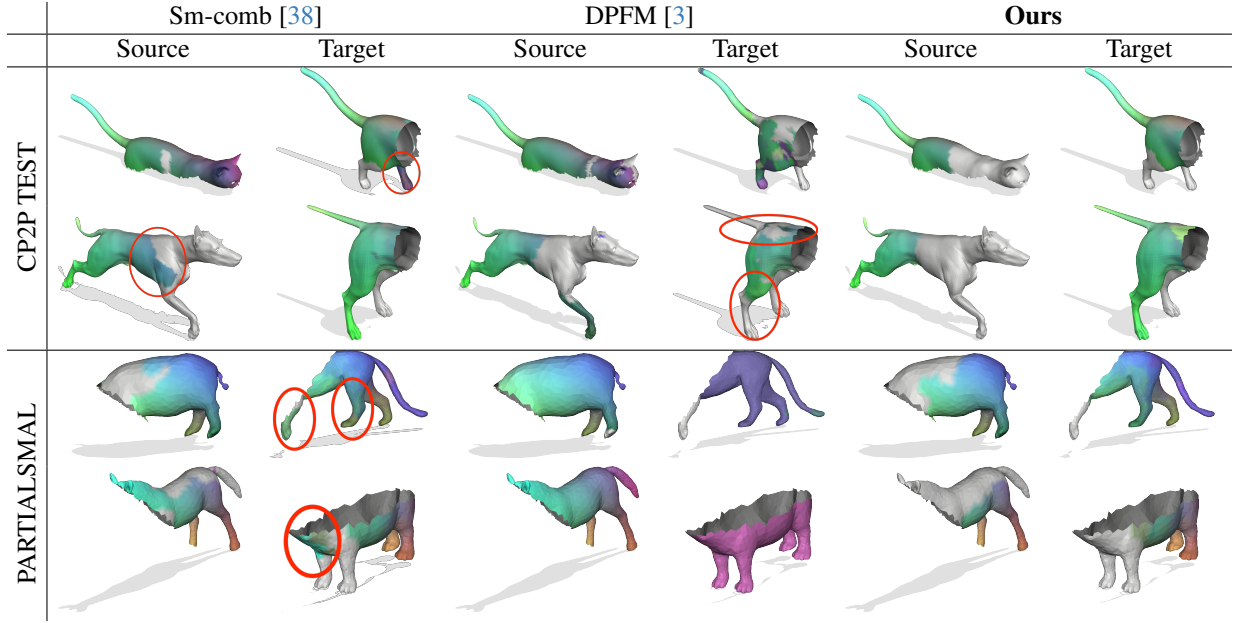

    \setlength\tabcolsep{4pt}
    \centering
        \begin{tabular}{c | c c c c c c}
             &\multicolumn{2}{c}{Sm-comb~\cite{roetzer2022scalable}}& \multicolumn{2}{c}{DPFM~\cite{attaiki2021dpfm}} &\multicolumn{2}{c}{\textbf{Ours}}\\
             \hline
             &Source & Target &Source & Target &Source & Target \\
             \hline
\multirow{2}{*}[1.2em]{\rotatebox{90}{CP2P TEST}}&\includeGeoErrorRowCat{cuts_cat_shape_6}{cuts_cat_shape_7}{0.14}{0.11}{vis/cat_geo_error} \\
&\includeGeoErrorRowCat{cuts_dog_shape_1}{cuts_dog_shape_12}{0.14}{0.11}{vis/dog_geo_error}\\
\hline
\multirow{2}{*}[1.2em]{\rotatebox{90}{PARTIALSMAL}}&\includeGeoErrorRowSmalNew{cuts_21_cougar_04}{cuts_23_hippo_06}{0.12}{0.12}{vis/geo_error_cougar_hippo}\\
&\includeGeoErrorRowSmalNew{cuts_2_hippo_05}{cuts_23_horse_01}{0.14}{0.10}{vis/geo_error_horse_hippo}
        \end{tabular}
        \vspace{-3mm}
    \caption{\textbf{Qualitative comparison} via \textbf{color transfer}. The results are shown on the predicted overlapping region of the methods. The geometric consistent matchings produced by our method visually improve matchings drastically compared to DPFM~\cite{attaiki2021dpfm} and Sm-comb~\cite{roetzer2022scalable}.}
    \label{fig:geo_error_qual}
\end{figure*}

\subsection{Correspondence Quality}

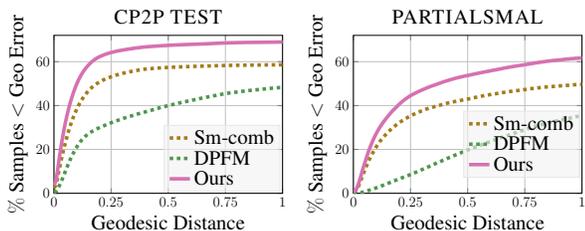
\begin{figure}[h]%
    \setlength\tabcolsep{-5pt} 
    \begin{tabular}{ll}
      {\newcommand{\pckLineWidth}{3pt}
\newcommand{\plotWidth}{1*\linewidth}
\newcommand{\plotHeight}{0.75*\linewidth}
\newcommand{\pckTitle}{\textsc{CP2P TEST}}

\pgfplotsset{%
    label style = {font=\LARGE},
    tick label style = {font=\large},
    title style =  {font=\LARGE},
    legend style={  fill= gray!10,
                    fill opacity=0.6, 
                    font=\LARGE,
                    draw=gray!20, 
                    text opacity=1}
}
\begin{tikzpicture}[scale=0.45, transform shape]
	\begin{axis}[
		width=\plotWidth,
		height=\plotHeight,
		grid=major,
		title=\pckTitle,
		legend style={
			at={(0.99,0.01)},
			anchor=south east,
			legend columns=1},
		legend cell align={left},
	ylabel={{\LARGE
        $\%$ Samples $<$ Geo Error}},
        xmin=0,
        xmax=1,
        xlabel=\LARGE
        Geodesic Distance,
        ylabel near ticks,
        xtick={0, 0.25, 0.5, 0.75, 1},
        xticklabels={$0$, $0.25$, $0.5$, $0.75$, $1$},
        ymin=0,
        ymax=72,
        ytick={0, 20, 40, 60},
        yticklabels={$0$, $20$, $40$, $60$},
	]
 \addplot [color=cPLOTroetzer, dashed, smooth, line width=\pckLineWidth]
    table[row sep=crcr]{%
0.0  2.485186451300459\\
0.01  4.3112412044554596\\
0.02  9.32220892801185\\
0.03  14.349913112839358\\
0.04  18.946386348745122\\
0.05  23.144389368430048\\
0.06  26.953536734752014\\
0.07  30.37881377660029\\
0.08  33.39671822921118\\
0.09  36.06956670370054\\
0.1  38.429051648007295\\
0.11  40.52823120531009\\
0.12  42.35535424322707\\
0.13  43.94104492493519\\
0.14  45.3625758482181\\
0.15  46.610688545138586\\
0.16  47.67897330712474\\
0.17  48.62974674529242\\
0.18  49.478677036150756\\
0.19  50.179115745093014\\
0.2  50.79444776799704\\
0.21  51.377375153120816\\
0.22  51.8823177506196\\
0.23  52.356992279862126\\
0.24  52.728755377033295\\
0.25  53.115830555792954\\
0.26  53.46943281201037\\
0.27  53.81021565108395\\
0.28  54.10933538444007\\
0.29  54.39563570065237\\
0.3  54.66804831495884\\
0.31  54.915890379739615\\
0.32  55.14877645785261\\
0.33  55.37525282739367\\
0.34  55.58107569153634\\
0.35  55.785474175996356\\
0.36  55.97135572458194\\
0.37  56.16151041221548\\
0.38  56.334216448736576\\
0.39  56.499088397003106\\
0.4  56.61802410050423\\
0.41  56.733398854798736\\
0.42  56.825627439250205\\
0.43  56.900407372589235\\
0.44  56.972694641483635\\
0.45  57.057445232601204\\
0.46  57.12047403355839\\
0.47  57.191693017690795\\
0.48  57.256146198330626\\
0.49  57.32166766373245\\
0.5  57.382559895165656\\
0.51  57.42600347548642\\
0.52  57.47621285929978\\
0.53  57.51467111073127\\
0.54  57.56203173517933\\
0.55  57.61437768851665\\
0.56  57.665299262171324\\
0.57  57.70055265931687\\
0.58  57.74185967011367\\
0.59  57.784234965672454\\
0.6  57.81663960345269\\
0.61  57.85438566504288\\
0.62  57.881448879013185\\
0.63  57.907087713300854\\
0.64  57.93521921203316\\
0.65  57.96584337521009\\
0.66  57.99931629775232\\
0.67  58.02994046092926\\
0.68  58.06412557331282\\
0.69  58.095818021251745\\
0.7  58.12857875395265\\
0.71  58.15884682220893\\
0.72  58.18982708030652\\
0.73  58.21546591459419\\
0.74  58.2492949320571\\
0.75  58.27742643078939\\
0.76  58.30377745491838\\
0.77  58.320157821268836\\
0.78  58.34259180127055\\
0.79  58.35363074381107\\
0.8  58.36431359143093\\
0.81  58.37642081873344\\
0.82  58.38817195111529\\
0.83  58.39564994444919\\
0.84  58.414879070164936\\
0.85  58.42200096857818\\
0.86  58.43197162635673\\
0.87  58.444078853659235\\
0.88  58.457610460644396\\
0.89  58.46864940318491\\
0.9  58.479332250804774\\
0.91  58.495000427313904\\
0.92  58.508175939378404\\
0.93  58.525980685411504\\
0.94  58.534170868586735\\
0.95  58.54022448223799\\
0.96  58.549482950175204\\
0.97  58.56123408255704\\
0.98  58.565507221604996\\
0.99  58.57369740478022\\
1  58.57369740478022\\
    };
    \addlegendentry{\textcolor{black}{Sm-comb}}
    \addplot [color=cPLOTdpfm, dashed, smooth, line width=\pckLineWidth]
    table[row sep=crcr]{%
0.0  0.5428125191400748\\
0.01  0.9497305077479022\\
0.02  2.729757456973112\\
0.03  4.933162859067801\\
0.04  7.408357322226985\\
0.05  10.055812457891836\\
0.06  12.655800208244013\\
0.07  15.130229068414284\\
0.08  17.340142095914743\\
0.09  19.321905432718808\\
0.1  21.13906412690635\\
0.11  22.736877564770015\\
0.12  24.088166840203346\\
0.13  25.27370306853678\\
0.14  26.220753965823484\\
0.15  27.05526122373982\\
0.16  27.787560482636124\\
0.17  28.393535248361612\\
0.18  28.959315857169106\\
0.19  29.4894959269921\\
0.2  29.979099038402644\\
0.21  30.435015618300977\\
0.22  30.857245666687085\\
0.23  31.293256568873645\\
0.24  31.728884669565748\\
0.25  32.140779077601515\\
0.26  32.54846266919826\\
0.27  32.917100508360384\\
0.28  33.28994916396154\\
0.29  33.64021253138972\\
0.3  34.0226312243523\\
0.31  34.370214981319286\\
0.32  34.70018986954125\\
0.33  35.00375145464567\\
0.34  35.3180314815949\\
0.35  35.65298278924481\\
0.36  35.96879402217186\\
0.37  36.26737918784835\\
0.38  36.56022233110798\\
0.39  36.8362222086115\\
0.4  37.12370613094873\\
0.41  37.42229129662522\\
0.42  37.691783548722974\\
0.43  37.97850186807129\\
0.44  38.264454584430695\\
0.45  38.57031297850187\\
0.46  38.88229619648435\\
0.47  39.18126416365529\\
0.48  39.46147485759784\\
0.49  39.707999020028176\\
0.5  39.957968395908615\\
0.51  40.20717216880015\\
0.52  40.47513321492007\\
0.53  40.72739939976726\\
0.54  40.96818153978073\\
0.55  41.210112084277576\\
0.56  41.45969865866356\\
0.57  41.674067495559505\\
0.58  41.88805353096098\\
0.59  42.10395357383475\\
0.6  42.3458841183316\\
0.61  42.55336252832731\\
0.62  42.770028174189996\\
0.63  42.99052183499724\\
0.64  43.23206957799963\\
0.65  43.43648557603969\\
0.66  43.647026397991056\\
0.67  43.8667544558094\\
0.68  44.07385006431065\\
0.69  44.28477368775648\\
0.7  44.482682060390765\\
0.71  44.6775280210694\\
0.72  44.88232682060391\\
0.73  45.082914803699396\\
0.74  45.262831506094194\\
0.75  45.441599804005634\\
0.76  45.613094873522385\\
0.77  45.763535860844\\
0.78  45.90019599436516\\
0.79  46.020012862130216\\
0.8  46.15322778220126\\
0.81  46.295247136644825\\
0.82  46.428844858210326\\
0.83  46.55670055735898\\
0.84  46.677283028112946\\
0.85  46.805904330250506\\
0.86  46.91806516812642\\
0.87  47.02563238806884\\
0.88  47.122863967660926\\
0.89  47.229282783119984\\
0.9  47.33417039260121\\
0.91  47.44403442151038\\
0.92  47.551984442947266\\
0.93  47.660700067373064\\
0.94  47.774774912721256\\
0.95  47.862053653457465\\
0.96  47.961199240521836\\
0.97  48.045032767807925\\
0.98  48.122358669688246\\
0.99  48.20198138053531\\
1  48.20198138053531\\
    };
    \addlegendentry{\textcolor{black}{DPFM}}
    \addplot [color=cPLOTours, smooth, line width=\pckLineWidth]
    table[row sep=crcr]{%
0.0  3.6808811362382796\\
0.01  6.3413541092112515\\
0.02  13.64925193050193\\
0.03  20.68653474903475\\
0.04  26.82880584666299\\
0.05  32.22645477109763\\
0.06  36.79329840044126\\
0.07  40.807708218422505\\
0.08  44.27140788747931\\
0.09  47.28781715388858\\
0.1  49.9267443463872\\
0.11  52.203702426916706\\
0.12  54.107487589630445\\
0.13  55.72514478764479\\
0.14  57.145011720904584\\
0.15  58.34812810259239\\
0.16  59.39783852730282\\
0.17  60.34369829012686\\
0.18  61.11417539988968\\
0.19  61.745466767788194\\
0.2  62.29229867622724\\
0.21  62.74820739106454\\
0.22  63.17179743519029\\
0.23  63.55445049641478\\
0.24  63.87246621621622\\
0.25  64.16333425261996\\
0.26  64.43222559293987\\
0.27  64.67914023717594\\
0.28  64.91571290678434\\
0.29  65.11479591836735\\
0.3  65.30052054605626\\
0.31  65.47331770546056\\
0.32  65.64051296194154\\
0.33  65.80253723110866\\
0.34  65.94861762272477\\
0.35  66.11150372311087\\
0.36  66.25456770546056\\
0.37  66.38901337562052\\
0.38  66.51268615554329\\
0.39  66.63032611693326\\
0.4  66.7419332597904\\
0.41  66.82897821290679\\
0.42  66.91214492553779\\
0.43  66.97936776061776\\
0.44  67.05736348593491\\
0.45  67.1241554054054\\
0.46  67.19654922779922\\
0.47  67.26635755653612\\
0.48  67.3275475730833\\
0.49  67.38227385548814\\
0.5  67.4533749310535\\
0.51  67.50982487589631\\
0.52  67.5598110865968\\
0.53  67.60376447876449\\
0.54  67.64857970215114\\
0.55  67.69339492553779\\
0.56  67.74725937672366\\
0.57  67.78302537231109\\
0.58  67.8261169332598\\
0.59  67.86834666298952\\
0.6  67.89807984004412\\
0.61  67.92781301709873\\
0.62  67.96875\\
0.63  68.00236141753999\\
0.64  68.04674572531717\\
0.65  68.08682087699944\\
0.66  68.13982349696636\\
0.67  68.18937879205737\\
0.68  68.22255929398786\\
0.69  68.27082184225041\\
0.7  68.32511720904579\\
0.71  68.37294884169884\\
0.72  68.42595146166575\\
0.73  68.46818119139547\\
0.74  68.5207528957529\\
0.75  68.56082804743518\\
0.76  68.59271580253723\\
0.77  68.61900165471594\\
0.78  68.63753102592388\\
0.79  68.65907680639823\\
0.8  68.6901027302813\\
0.81  68.71035576392718\\
0.82  68.72802330391616\\
0.83  68.74396718146718\\
0.84  68.76766753998896\\
0.85  68.78274958632102\\
0.86  68.79481522338665\\
0.87  68.80860452289023\\
0.88  68.8133445945946\\
0.89  68.82411748483177\\
0.9  68.833166712631\\
0.91  68.84566326530613\\
0.92  68.85126516822946\\
0.93  68.86591629895202\\
0.94  68.87927468284612\\
0.95  68.88875482625483\\
0.96  68.899527716492\\
0.97  68.91590250965251\\
0.98  68.94434293987865\\
0.99  68.96890512961942\\
1  68.96890512961942\\
    };
    \addlegendentry{\textcolor{black}{Ours}}
	\end{axis}
\end{tikzpicture}}   & \newcommand{\pckLineWidth}{3pt}
\newcommand{\plotWidth}{1*\linewidth}
\newcommand{\plotHeight}{0.75*\linewidth}
\newcommand{\pckTitle}{\textsc{PARTIALSMAL}}

\pgfplotsset{%
    label style = {font=\LARGE},
    tick label style = {font=\large},
    title style =  {font=\LARGE},
    legend style={  fill= gray!10,
                    fill opacity=0.6, 
                    font=\LARGE,
                    draw=gray!20, 
                    text opacity=1}
}
\begin{tikzpicture}[scale=0.45, transform shape]
	\begin{axis}[
		width=\plotWidth,
		height=\plotHeight,
		grid=major,
		title=\pckTitle,
		legend style={
			at={(0.99,0.1)},
			anchor=south east,
			legend columns=1},
		legend cell align={left},
	ylabel={{\LARGE
        $\%$ Samples $<$ Geo Error}},
        xmin=0,
        xmax=1,
        xlabel=\LARGE
        Geodesic Distance,
        ylabel near ticks,
        xtick={0, 0.25, 0.5, 0.75, 1},
        xticklabels={$0$, $0.25$, $0.5$, $0.75$, $1$},
        ymin=0,
        ymax=72,
        ytick={0, 20, 40, 60},
        yticklabels={$0$, $20$, $40$, $60$},
	]
 \addplot [color=cPLOTroetzer, dashed, smooth, line width=\pckLineWidth]
    table[row sep=crcr]{%
0.0  0.7355486761629045\\
0.01  1.01953268073876\\
0.02  2.7043706442824385\\
0.03  5.130276407752864\\
0.04  7.9048904453932884\\
0.05  10.571630700376804\\
0.06  13.093368521928483\\
0.07  15.444515245325052\\
0.08  17.482075391230616\\
0.09  19.407246107763402\\
0.1  21.12319185979419\\
0.11  22.752838585699447\\
0.12  24.14716994385546\\
0.13  25.41305623987115\\
0.14  26.598664372049154\\
0.15  27.720551711680976\\
0.16  28.714997466220453\\
0.17  29.717471037143707\\
0.18  30.522761368140607\\
0.19  31.369194258104333\\
0.2  32.12481247522666\\
0.21  32.84229856452608\\
0.22  33.49807081537527\\
0.23  34.127752662977215\\
0.24  34.72783193766401\\
0.25  35.342963368070365\\
0.26  35.82864625928843\\
0.27  36.27820397677954\\
0.28  36.7573642671858\\
0.29  37.18584896666951\\
0.3  37.622863221060975\\
0.31  38.03930619596903\\
0.32  38.42815355205788\\
0.33  38.75026968445664\\
0.34  39.04027455132032\\
0.35  39.326767248516106\\
0.36  39.615266899807835\\
0.37  39.909285674863405\\
0.38  40.20480966549093\\
0.39  40.488793670066784\\
0.4  40.77679158283452\\
0.41  41.073319050510015\\
0.42  41.30512224859137\\
0.43  41.51484895161735\\
0.44  41.74163476445885\\
0.45  41.96691536172839\\
0.46  42.175136849182415\\
0.47  42.36429227272499\\
0.48  42.56699463641518\\
0.49  42.77621960091717\\
0.5  42.9603576392199\\
0.51  43.170586080769866\\
0.52  43.402389278851224\\
0.53  43.627669876120756\\
0.54  43.84141048733863\\
0.55  44.077227593611866\\
0.56  44.26989518682234\\
0.57  44.4700888578926\\
0.58  44.66978079043887\\
0.59  44.825319732874405\\
0.6  45.0245099268967\\
0.61  45.188076685716005\\
0.62  45.34311388962756\\
0.63  45.52524497383433\\
0.64  45.673759576934074\\
0.65  45.84836458328107\\
0.66  46.03099740601183\\
0.67  46.18452939435143\\
0.68  46.33806138269102\\
0.69  46.48055512350294\\
0.7  46.61803147907499\\
0.71  46.74898523383524\\
0.72  46.8689007410678\\
0.73  47.00035623435203\\
0.74  47.1338186817322\\
0.75  47.29036110121571\\
0.76  47.454429598559\\
0.77  47.60194072461078\\
0.78  47.73138926379906\\
0.79  47.86133954151134\\
0.8  47.98677417250774\\
0.81  48.086118400256886\\
0.82  48.20904433863337\\
0.83  48.32444419915006\\
0.84  48.42479190394718\\
0.85  48.53467264070003\\
0.86  48.61745949715765\\
0.87  48.68770289051564\\
0.88  48.77550713221312\\
0.89  48.85879572719473\\
0.9  48.941582583652355\\
0.91  49.01483640815425\\
0.92  49.092104140848036\\
0.93  49.16736491944588\\
0.94  49.24914829885553\\
0.95  49.33895949464896\\
0.96  49.42074287405862\\
0.97  49.51155754690001\\
0.98  49.59133397221372\\
0.99  49.66207910409569\\
1  49.66207910409569\\
    };
    \addlegendentry{\textcolor{black}{Sm-comb}}
    \addplot [color=cPLOTdpfm, dashed, smooth, line width=\pckLineWidth]
    table[row sep=crcr]{%
0.0  0.030009790079058577\\
0.01  0.035421391568724876\\
0.02  0.12397487049053707\\
0.03  0.2735318571140421\\
0.04  0.4757289673188466\\
0.05  0.7669715202172512\\
0.06  1.0803524428461089\\
0.07  1.410952097487541\\
0.08  1.742535679672549\\
0.09  2.112000472285221\\
0.1  2.471625989462136\\
0.11  2.873560391012806\\
0.12  3.262703734496992\\
0.13  3.647419404035087\\
0.14  4.094614472590238\\
0.15  4.489661381335878\\
0.16  4.888152036484033\\
0.17  5.305337314960126\\
0.18  5.713667245544038\\
0.19  6.1421676907712515\\
0.2  6.549513693811587\\
0.21  6.9548918417647725\\
0.22  7.321896815518506\\
0.23  7.726291035928114\\
0.24  8.114450451868723\\
0.25  8.548362498585604\\
0.26  8.979322762671757\\
0.27  9.42405801236797\\
0.28  9.897819124599664\\
0.29  10.326319569826877\\
0.3  10.786797660220302\\
0.31  11.24137218535227\\
0.32  11.665936920405182\\
0.33  12.146093561670119\\
0.34  12.646912681350145\\
0.35  13.124117539984356\\
0.36  13.569836717224144\\
0.37  14.08000314856814\\
0.38  14.491284861782777\\
0.39  14.913881741748538\\
0.4  15.346317897150055\\
0.41  15.832870067448232\\
0.42  16.295316012928808\\
0.43  16.709549508774174\\
0.44  17.14887315698072\\
0.45  17.606891428515205\\
0.46  17.99603477199939\\
0.47  18.45454500730566\\
0.48  18.897312401914725\\
0.49  19.30465840495506\\
0.5  19.711512444223608\\
0.51  20.128205758927912\\
0.52  20.538011580827188\\
0.53  20.922727250365284\\
0.54  21.29760364446762\\
0.55  21.712821067856563\\
0.56  22.071954621261693\\
0.57  22.499471138945328\\
0.58  22.858112728578668\\
0.59  23.23643286908352\\
0.6  23.65460207510319\\
0.61  24.03833381709771\\
0.62  24.395991479187472\\
0.63  24.729050952687846\\
0.64  25.10491127433376\\
0.65  25.48126355975146\\
0.66  25.813339105708256\\
0.67  26.156729818416174\\
0.68  26.521766937082752\\
0.69  26.90402278776191\\
0.7  27.24888939178519\\
0.71  27.61392651045177\\
0.72  27.976503810259413\\
0.73  28.30119989963939\\
0.74  28.640654902173\\
0.75  28.970762593042647\\
0.76  29.310217595576262\\
0.77  29.65213241696881\\
0.78  30.00241062248176\\
0.79  30.359576320799736\\
0.8  30.679844736233626\\
0.81  30.92631858589933\\
0.82  31.213625428623438\\
0.83  31.487157285737478\\
0.84  31.75822932399258\\
0.85  32.042584384085956\\
0.86  32.282662704718426\\
0.87  32.52470888043805\\
0.88  32.74756846905794\\
0.89  32.991582499864705\\
0.9  33.198699247787395\\
0.91  33.426970437896955\\
0.92  33.666064830985846\\
0.93  33.912046716879765\\
0.94  34.153108965055814\\
0.95  34.3557980390324\\
0.96  34.585545120457326\\
0.97  34.80840470907722\\
0.98  35.03913571804572\\
0.99  35.25953548780667\\
1  35.25953548780667\\
    };
    \addlegendentry{\textcolor{black}{DPFM}}
    \addplot [color=cPLOTours, smooth, line width=\pckLineWidth]
    table[row sep=crcr]{%
0.0  0.939331015854103\\
0.01  1.3598320309562781\\
0.02  3.6856017444718865\\
0.03  6.899802763622903\\
0.04  10.553592189208105\\
0.05  13.9349524839637\\
0.06  17.145683068327077\\
0.07  20.18057829011678\\
0.08  22.744072786585612\\
0.09  25.16007380457982\\
0.1  27.20705192348848\\
0.11  29.096125259559603\\
0.12  30.7839133779477\\
0.13  32.27909236562188\\
0.14  33.66321744009162\\
0.15  34.99181555795916\\
0.16  36.21167338581402\\
0.17  37.436158460052404\\
0.18  38.45068222963867\\
0.19  39.49181266593016\\
0.2  40.52715904424226\\
0.21  41.48789107462013\\
0.22  42.32310904684509\\
0.23  43.11668180161838\\
0.24  43.846629918618305\\
0.25  44.50312049927989\\
0.26  45.10003528275367\\
0.27  45.60787557334475\\
0.28  46.14694977702457\\
0.29  46.61545847335574\\
0.3  47.079339923303394\\
0.31  47.544956590644865\\
0.32  47.98628021447287\\
0.33  48.39984035999977\\
0.34  48.80819485334521\\
0.35  49.163336013280194\\
0.36  49.517320361619305\\
0.37  49.89791137666364\\
0.38  50.28486485548531\\
0.39  50.65793659515643\\
0.4  50.997460798547046\\
0.41  51.31674079900977\\
0.42  51.657421813996265\\
0.43  51.95703601732904\\
0.44  52.24913094528859\\
0.45  52.5128839891491\\
0.46  52.78010746779726\\
0.47  53.03518442468867\\
0.48  53.275222830833656\\
0.49  53.49790906304045\\
0.5  53.709605585086386\\
0.51  53.96005529559428\\
0.52  54.16249732487318\\
0.53  54.38286993388822\\
0.54  54.605556166095006\\
0.55  54.820144717130646\\
0.56  55.02952761598482\\
0.57  55.22329355829463\\
0.58  55.44540138470349\\
0.59  55.62875602265037\\
0.6  55.8491286316654\\
0.61  56.0602467479134\\
0.62  56.246493414849986\\
0.63  56.44257298035156\\
0.64  56.65369109659956\\
0.65  56.85266269109081\\
0.66  57.089230662448166\\
0.67  57.277212546778564\\
0.68  57.45073428616049\\
0.69  57.65491153283321\\
0.7  57.81165950407486\\
0.71  57.9869164608506\\
0.72  58.1587029828387\\
0.73  58.309666896100964\\
0.74  58.45773878037354\\
0.75  58.63588776613897\\
0.76  58.79205733158269\\
0.77  58.9626870419749\\
0.78  59.135051969760944\\
0.79  59.31956341930371\\
0.8  59.47457617315156\\
0.81  59.639421825564376\\
0.82  59.79385617361428\\
0.83  59.93556559410952\\
0.84  60.079010231998566\\
0.85  60.21435718871646\\
0.86  60.330038348304406\\
0.87  60.49083516013164\\
0.88  60.583958493599944\\
0.89  60.67881704446205\\
0.9  60.78755733447472\\
0.91  60.884151102730655\\
0.92  61.007929943489756\\
0.93  61.13113037845091\\
0.94  61.22656733511097\\
0.95  61.33646443671952\\
0.96  61.41454921944137\\
0.97  61.50304530652615\\
0.98  61.5707187848851\\
0.99  61.62219690090174\\
1  61.62219690090174\\
    };
    \addlegendentry{\textcolor{black}{Ours}}
	\end{axis}
\end{tikzpicture}
    \end{tabular}
\caption{
Comparison of \textbf{geodesic errors} on the \textsc{CP2P TEST} dataset (left) and on the PARTIALSMAL dataset (right).
Our method outperforms current SOTA
methods~\cite{attaiki2021dpfm, roetzer2022scalable}.
}
\label{fig:geo_errors}
\end{figure}

\paragraph{Geodesic Error}
We assess the geodesic error of the correspondences according to the Princeton Protocol~\cite{kim2011blended}.
This assessment involves normalizing the geodesic error between the ground truth and computed correspondences by the shape diameter, which is equivalent to the square root of the area of the target shape.
Ours and Sm-comb~\cite{roetzer2022scalable} establish \emph{triangle-to-triangle} correspondences, including cases with degenerate triangles, possibly resulting in non-bijective
correspondences between vertices.
To compare with DPFM~\cite{attaiki2021dpfm} that returns vertex-wise point-to-point maps, we transform our \emph{triangle-to-triangle} to 
vertex correspondences.
In the DPFM~\cite{attaiki2021dpfm} paper the geodesic error is compared only on the ground truth overlapping region.
To better account for the partial-to-partial matching paradigm, we set
all vertices on  the shape that remain unmatched
to an infinite geodesic error, and compute the geodesic error for the remaining vertices, such that we obtain a metric, that includes the overall performance of the matching.
Therefore, the curve is expected to be lower than a comparable full-to-full matching.
\paragraph{Results} 
We show that our method outperforms DPFM~\cite{attaiki2021dpfm} and Sm-comb~\cite{roetzer2022scalable} in terms of geodesic error on the CP2P TEST dataset and {the PARTIALSMAL dataset }(see Figure \ref{fig:geo_errors}).
We show qualitative color transfer examples in Figure~\ref{fig:geo_error_qual}.

\subsection{Faster Optimization}
As ablation we evaluate our heuristic from Section~\ref{sec:heuristic} on the CP2P Test dataset. 
\begin{table}
\begin{tabular}{l|c|c|c|c}
    \toprule
    Limit & Time (min) & mIoU &$\text{AUC}_{\text{GeoErr}}$ & GlobOpt\\
    \hline
    15min & 
    {299.55}
    & 
    {\textbf{69.29}}
    & 
    {\textbf{61.25}}
    &
    {$\mathbf{22\%}$}
    \\
    5min & 
    {140.13}
    & 
    {69.15}
    & 
    {60.38}
    &
    {$10\%$}
    \\
    1min & 
    {57.31}
    & 
    {69.47}
    & 
    {60.96}
    &
    {$8\%$}
    \\
    10s &
    {\textbf{50.68}}
    &
    {68.89}
    &
    {60.95}
    &{$4\%$}\\
    \bottomrule
\end{tabular}%
    \caption{
    We \textbf{reduce the time limit} for subproblems of the minimum mean algorithm and get similar results regarding mean IoU and $\text{AUC}_{\text{GeoErr}}$, but ensure global optimality in fewer cases.}
    \label{tab:time_limit_vs_glob_opt}
\end{table}
In Table~\ref{tab:time_limit_vs_glob_opt} we show that we can get faster  optimization while achieving similar results w.r.t.~mean IoU and $\text{AUC}_{\text{GeoErr}}$. Yet,this comes at the cost of running more frequently 
into time limits for the linear subproblems and thus we can ensure global optimality (on the coarse level) in fewer cases.
We show the optimization time for the individual values of $x$ in \eqref{eq:ilp} in the supplementary material.

\section{Discussion and Limitations}
While our method outperforms current SOTA methods, there are some limitations.
Even though our method uses a pruned search algorithm to reduce runtime, runtimes for solving individual ILPs are still high.
Another drawback is that our neighboring constraint is only defined on the inner triangles of the matching, so that there are no rules for matching a boundary edge on a triangle.
Thus, triangles can be matched to boundaries, even though their neighborhood is not matched.
Additionally, due to our coarse-to-fine scheme, we only guarantee geometric consistency at coarse levels, but not at finer levels.
Further, our approach is based on explicit shape representations. Transferring analogous concepts to implicit representations~\cite{zheng2021deep} is an interesting direction for future work.
\section{Conclusion}
Our paper introduces an innovative integer non-linear approach for partial-to-partial shape matching by exploiting geometric consistency as a constraint.
Our method outperforms existing approaches, showcasing SOTA results in terms of intersection over union and geodesic error. 
This research represents a significant stride forward in addressing real-world challenges in partial-to-partial shape matching, e.g.~from 3D scanning. 
It differs from previous work by eliminating the need for annotated data or closing shapes. 
Despite potential challenges, such as high optimization time, our work provides a valuable tool for partial-to-partial shape matching.
We hope to encourage further research in the practically relevant area of partial-to-partial shape matching utilizing geometric consistency.
\section*{Acknowledgments}
We thank Daniel Scholz for providing feedback on the manuscript and enhancing our figures. 
We thank Yizheng Xie for help regarding evaluating DPFM.
This work was supported by the CRC \textit{Discretization in Geometry and Dynamics}.

{
     \small
     \bibliographystyle{ieeenat_fullname}
     \bibliography{main}
}

\newcommand{\psVis}[3]{ \includegraphics[width=0.08\linewidth]{vis/supp_vis/cuts_#1_#2_#3_cougar_01.png}
}

\clearpage
\setcounter{page}{1}
\twocolumn[{%
\renewcommand\twocolumn[1][]{#1}%
\maketitlesupplementary
\begin{center}%
\normalsize
    \centering%
    \captionsetup{type=figure}%
    \begin{tabular}{c | c c c | c c c | c c c}
    &\multicolumn{3}{c}{$z=-1$} & \multicolumn{3}{c}{$z=0$} &
    \multicolumn{3}{c}{$z=1$}\\
    \hline
&$x=-1$&$x=0$&$x=1$&$x=-1$&$x=0$&$x=1$&$x=-1$&$x=0$&$x=1$\\
\hline
    \rotatebox{90}{$y=-1$}&\psVis{-1}{-1}{-1} & \psVis{0}{-1}{-1}& \psVis{1}{-1}{-1}&\psVis{-1}{-1}{0}&\psVis{0}{-1}{0}&\psVis{1}{-1}{0}&\psVis{-1}{-1}{1}&\psVis{0}{-1}{1}&\psVis{1}{-1}{1}\\
    \rotatebox{90}{$y=0$}&\psVis{-1}{0}{-1} & \psVis{0}{0}{-1}& \psVis{1}{0}{-1}&\psVis{-1}{0}{0}&&\psVis{1}{0}{0}&\psVis{-1}{0}{1}&\psVis{0}{0}{1}&\psVis{1}{0}{1}\\
    \rotatebox{90}{$y=1$}&\psVis{-1}{1}{-1} & \psVis{0}{1}{-1}& \psVis{1}{1}{-1}&\psVis{-1}{1}{0}&\psVis{0}{1}{0}&\psVis{1}{1}{0}&\psVis{-1}{1}{1}&\psVis{0}{1}{1}&\psVis{1}{1}{1}\\
    \end{tabular}
    \caption{
        We generate our \textbf{PARTIALSMAL dataset} from shapes of the SMAL dataset~\cite{Zuffi:CVPR:2017} by rotating a plane with normal vector $(x,y,z)$ 
        around the origin and cutting off one half of the shape (as shown in gray).
        We show the resulting 26 generated partial shapes (red) of an exemplary cougar shape.
        The shown values $x,y,z$ in the table correspond to the normal vector $(x,y,z)$.
    }
    \label{fig:suppPartialSmal}
\end{center}
}]
\section{Data Preprocessing}
\label{sec:daraPrep}
\paragraph{Orientation and Manifoldness.}
Our algorithm requires all shapes in our datasets to be edge-manifold, vertex-manifold, and oriented.
We consider a shape to be oriented if all triangle normals point outwards which we ensure for every shape.
We achieve vertex manifoldness by duplicating each non-manifold vertex, and use the \emph{libigl}\footnote{\url{https://libigl.github.io}} library to achieve edge-manifoldness.
Furthermore, we remove all disconnected shape parts.

\paragraph{Shape Decimation.}
The complexity of our Integer Linear Program increases quadratically with the number of triangles of respective shapes $\mathcal{X}$ and $\mathcal{Y}$.
For manageable runtimes and memory consumption when solving our Integer Program (\cref{eq:ilp}), we reduce the amount of triangles of shapes $\mathcal{X}$ and $\mathcal{Y}$.
To achieve reduction in the number of triangles, we employ a straightforward edge collapse method, where two vertices are merged into one during each step of the reduction process.
We apply these reduction steps until a predefined number of vertices is attained.
We reduce a partial shape $\mathcal{X}$ proportionally to its surface area $A(\cdot)$ to $100 \cdot A(\mathcal{X})$ number of triangles.
\paragraph{Feature Computation.}
We compute features $\mathbf{W}^{(\mathcal{X})}$
on the full-resolution shape $\mathcal{X}$ and transfer them to the low-resolution counterpart.
This involves associating each low-resolution vertex with a high-resolution vertex using the edge-collapse algorithm.
Additionally, we compute Voronoi areas for each vertex on the low-resolution shape.
We then average the features of all high-resolution vertices that fall within the Voronoi area around the corresponding mapped high-resolution vertex.

\section{PARTIALSMAL Dataset Generation}
The first step of generating the PARTIALSMAL dataset is mean-centering all shapes of the SMAL test dataset \cite{Zuffi:CVPR:2017}.
We then define planes with origin at $(0,0,0)$ and normals $(x,y,z)$, with $(x,y,z) \in \{-1,0,1\}^3 \backslash (0,0,0)$.
The shape is then cut along this plane, i.e., all triangles above the plane (all triangles lying on the side of the plane with positive normal direction) are discarded.
We consider the largest connected part of the remaining triangles as the resulting partial shape.
We show examples of generated shapes in Figure~\ref{fig:suppPartialSmal}.

\section{Ablation Studies}
{We show analysis in terms of runtime, performance and memory consumption. 
Additionally, we reason the choice of our objective function and compare our method to the comparison method in terms of overlapping regions.}
\subsection{Reduction of Optimization Time}
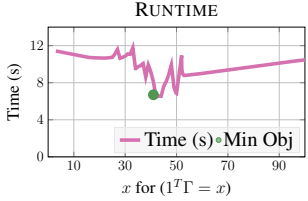
\begin{figure}
\centering
\newcommand{\pckLineWidth}{3pt}
\newcommand{\plotWidth}{1.1*\linewidth}
\newcommand{\plotHeight}{0.65*\linewidth}
\newcommand{\pckTitle}{\textsc{Runtime}}

\pgfplotsset{%
    label style = {font=\LARGE},
    tick label style = {font=\large},
    title style =  {font=\LARGE},
    legend style={  fill= gray!10,
                    fill opacity=0.6, 
                    font=\LARGE,
                    draw=gray!20, 
                    text opacity=1},
}
\begin{tikzpicture}[scale=0.45, transform shape]
	\begin{axis}[
		width=\plotWidth,
		height=\plotHeight,
		grid=major,
		title=\pckTitle,
		legend style={
			at={(0.98,0.03)},
			anchor=south east,
			legend columns=2},
		legend cell align={left},
	ylabel={\Large Time (s)},
        xmin=0,
        xmax=100,
        ymin=0,
        ymax=14,
        ylabel near ticks,
        xlabel near ticks,
        xlabel={\Large $x$ for ($1^T\Gamma = x$)},
        xtick={10,30,50,70,90},
        xticklabels={$10$, $30$, $50$, $70$, $90$},
        ytick={0,4,8,12},
        yticklabels={$0$, $4$, $8$,$12$},
	]
 \addplot [color=cPLOTours, smooth, tension=0.1, line width=\pckLineWidth]
    table[row sep=crcr]{%
3 11.423452 \\
16 10.721168 \\
22 10.655647 \\
25 10.739956 \\
26 11.102121 \\
27 11.590428 \\
28 10.640856 \\
30 10.865454 \\
31 11.104301 \\
32 11.065309 \\
33 11.783285 \\
34 9.531029 \\
35 9.729357 \\
36 9.83922 \\
37 10.072697 \\
38 8.626801 \\
39 9.9164 \\
40 9.145467 \\
41 8.096899 \\
42 6.511665 \\
43 6.55696 \\
44 6.540483 \\
45 7.607138 \\
46 8.139044 \\
47 8.899112 \\
48 9.930523 \\
49 7.528324 \\
50 6.853362 \\
52 10.935013 \\
53 8.776527 \\
103 10.566622 \\
153 10.613082 \\
    };
\node [cPLOTdpfm,fill,circle,scale=0.6] at (41,67) {\textbullet};
\addlegendimage{only marks, mark=*,cPLOTdpfm}
\addlegendentry{\textcolor{black}{Time (s)}}
\addlegendentry{Min Obj}
\end{axis}
\end{tikzpicture}
\caption{\textbf{Optimization Time} for linear subproblems of the minimum mean algorithm of one example shape w.r.t. number of binary variables, that equal 1. We observe that the minimal mean solution often needs the least optimization time.}
\label{fig:opt_vs_nr_elements}
\end{figure}

For a sample shape, we display the optimization time over $x$ (i.e.~number of binary elements equal to one) in Figure~\ref{fig:opt_vs_nr_elements}.

\subsection{Scalability}
\begin{figure}
{
\setlength\tabcolsep{-5pt} %
\begin{tabular}{cc}
\hspace{-0.5cm}
\newcommand{\pckLineWidth}{3pt}
\newcommand{\plotWidth}{1.1*\linewidth}
\newcommand{\plotHeight}{0.65*\linewidth}
\newcommand{\pckTitle}{\textsc{Runtime}}

\pgfplotsset{%
    label style = {font=\LARGE},
    tick label style = {font=\large},
    title style =  {font=\LARGE},
    legend style={  fill= gray!10,
                    fill opacity=0.6, 
                    font=\LARGE,
                    draw=gray!20, 
                    text opacity=1}
}
\begin{tikzpicture}[scale=0.45, transform shape]
	\begin{axis}[
		width=\plotWidth,
		height=\plotHeight,
		grid=major,
		title=\pckTitle,
		legend style={
			at={(0.01,0.56)},
			anchor=south west,
			legend columns=1},
		legend cell align={left},
	ylabel={\Large Time (h)},
        xmin=4000,
        xmax=13664,
        xlabel={\Large Faces $\times10^4$},
        xtick scale label code/.code={},
        ylabel near ticks,
        xtick={4000,6000,8000,10000,12000, 14000},
        xticklabels={$4$, $6$, $8$, $10$, $12$,$14$},
        ymin=0,
        ymax=10.5,
        ytick={0, 2, 4, 6, 8, 10},
        yticklabels={$0$, $2$, $4$, $6$, $8$,$10$},
	]
 \addplot [color=cPLOTours, smooth, tension=0.1, line width=\pckLineWidth]
    table[row sep=crcr]{%
4295.0 4.087432957777777 \\
4453.0 0.08930066083333334 \\
5069.5 2.8025114108333327 \\
5102.0 0.8290480369444445 \\
5196.0 3.9500320175 \\
5232.0 5.394824736111111 \\
5360.5 1.8198615105555556 \\
5376.5 4.4966917266666675 \\
5394.5 1.0967711341666668 \\
5458.0 7.087147546111111 \\
5899.0 10.0 \\
6126.0 10.0 \\
6128.5 7.1651843830555535 \\
6228.5 0.42230680555555555 \\
6454.0 1.0461505633333332 \\
6472.0 1.7304019588888888 \\
6530.5 2.3855871075 \\
6570.5 0.02289762055555555 \\
6609.5 0.02323439666666667 \\
6624.0 10.0 \\
6660.5 0.8884121222222219 \\
6750.5 0.0522921 \\
7049.5 3.406321815833334 \\
7189.0 4.590794542777777 \\
7327.5 8.967715573888889 \\
7365.0 1.1900575519444445 \\
7599.0 10.0 \\
7633.0 9.100213759722218 \\
7884.0 1.9046266947222221 \\
7929.5 5.011344271666668 \\
7943.0 10.0 \\
8727.5 6.75791374527778 \\
8912.5 3.3206402719444448 \\
8984.5 4.006895925555556 \\
9001.5 5.455166271944445 \\
9125.5 3.2073234666666672 \\
9161.5 3.6867054016666665 \\
9164.0 2.567326984444445 \\
9282.5 4.670168398611111 \\
9401.5 2.6718982736111108 \\
9446.5 5.90546650388889 \\
9740.0 3.7781085622222226 \\
10060.0 9.844194844722225 \\
10264.5 6.378612036944445 \\
10927.5 8.42808250361111 \\
10940.0 7.671581996111112 \\
10940.0 3.6273648049999982 \\
11889.0 4.779277150833334 \\
13599.0 5.008274110277778 \\
13664.0 4.8171953230555555 \\
    };
 \addplot [color=cPLOTdpfm, dotted, smooth, line width=\pckLineWidth]
    table[row sep=crcr]{%
4295.0 2.619745868728754 \\
4453.0 2.7516468780596917 \\
5069.5 3.23721248516141 \\
5102.0 3.2615245937860893 \\
5196.0 3.3311180254332413 \\
5232.0 3.357485618191303 \\
5360.5 3.4503150580934454 \\
5376.5 3.461732688539392 \\
5394.5 3.4745402259874254 \\
5458.0 3.51940697936518 \\
5899.0 3.8174425230159637 \\
6126.0 3.9616120336187235 \\
6128.5 3.9631648378058557 \\
6228.5 4.024652344772383 \\
6454.0 4.158833479007177 \\
6472.0 4.169277057282326 \\
6530.5 4.202945953799824 \\
6570.5 4.225727308094941 \\
6609.5 4.247751364722118 \\
6624.0 4.25589251989589 \\
6660.5 4.276272328130241 \\
6750.5 4.325830079833477 \\
7049.5 4.483383661913571 \\
7189.0 4.553163204584625 \\
7327.5 4.620096066801899 \\
7365.0 4.637816462494129 \\
7599.0 4.744519994304731 \\
7633.0 4.759468618311671 \\
7884.0 4.865465115719908 \\
7929.5 4.883857460338142 \\
7943.0 4.889265988927481 \\
8727.5 5.165409707002461 \\
8912.5 5.219599147934738 \\
8984.5 5.239561412685607 \\
9001.5 5.244182519209413 \\
9125.5 5.276823895364894 \\
9161.5 5.285949395185873 \\
9164.0 5.286577244736591 \\
9282.5 5.31546348955304 \\
9401.5 5.342749230145741 \\
9446.5 5.352617611771167 \\
9740.0 5.410926768596412 \\
10060.0 5.462536406779693 \\
10264.5 5.488981484507478 \\
10927.5 5.539666623705615 \\
10940.0 5.540107642730682 \\
10940.0 5.540107642730682 \\
11889.0 5.517982009724045 \\
13599.0 5.201015107194997 \\
13664.0 5.1819354179421895 \\
    };
\end{axis}
\end{tikzpicture}&
\hspace{0.3cm}
\newcommand{\pckLineWidth}{3pt}
\newcommand{\plotWidth}{1.1*\linewidth}
\newcommand{\plotHeight}{0.65*\linewidth}
\newcommand{\pckTitle}{\textsc{Memory}}

\pgfplotsset{%
    label style = {font=\LARGE},
    tick label style = {font=\large},
    title style =  {font=\LARGE},
    legend style={  fill= gray!10,
                    fill opacity=0.6, 
                    font=\LARGE,
                    draw=gray!20, 
                    text opacity=1}
}
\begin{tikzpicture}[scale=0.45, transform shape]
	\begin{axis}[
		width=\plotWidth,
		height=\plotHeight,
		grid=major,
		title=\pckTitle,
		legend style={
			at={(0.01,0.56)},
			anchor=south west,
			legend columns=1},
		legend cell align={left},
	ylabel={\Large Size Search Space $\times10^5$ 
 },
        xmin=4000,
        xmax=13664,
        xlabel={\Large Faces $\times10^4$},
        xtick scale label code/.code={},
        ylabel near ticks,
        xtick={4000,6000,8000,10000,12000, 14000},
        xticklabels={$4$, $6$, $8$, $10$, $12$,$14$},
        ymin=0,
        ymax=400000,
        ytick={0,100000, 200000, 300000, 400000},
        yticklabels={$0$, $1$,$2$,$3$,$4$},
        ytick scale label code/.code={},
	]
 \addplot [color=cPLOTours, smooth, tension=0.2,  line width=\pckLineWidth]
    table[row sep=crcr]{%
4295.0 30654 \\
4453.0 34697 \\
5069.5 70434 \\
5102.0 68908 \\
5196.0 64133 \\
5232.0 61250 \\
5360.5 82612 \\
5376.5 63978 \\
5394.5 80058 \\
5458.0 64292 \\
5899.0 63586 \\
6126.0 64950 \\
6128.5 63586 \\
6228.5 70544 \\
6454.0 72266 \\
6472.0 102312 \\
6530.5 81826 \\
6570.5 50134 \\
6609.5 50323 \\
6624.0 97215 \\
6660.5 106440 \\
6750.5 60724 \\
7049.5 137830 \\
7189.0 140851 \\
7327.5 138285 \\
7365.0 92399 \\
7599.0 121215 \\
7633.0 123909 \\
7884.0 119660 \\
7929.5 124024 \\
7943.0 160558 \\
8727.5 160262 \\
8912.5 173319 \\
8984.5 184978 \\
9001.5 175952 \\
9125.5 228514 \\
9161.5 233512 \\
9164.0 227690 \\
9282.5 245169 \\
9401.5 252874 \\
9446.5 257565 \\
9740.0 255657 \\
10060.0 179332 \\
10264.5 200892 \\
10927.5 235272 \\
10940.0 237136 \\
10940.0 239000 \\
11889.0 239700 \\
13599.0 302270 \\
13664.0 323870 \\
    };
 \addplot [color=cPLOTdpfm, dotted, smooth, line width=\pckLineWidth]
    table[row sep=crcr]{%
4295.0 15754.696356764907 \\
4453.0 21951.912946215132 \\
5069.5 45780.113388281694 \\
5102.0 47020.68254599915 \\
5196.0 50600.006451691675 \\
5232.0 51967.35426317103 \\
5360.5 56832.41161108401 \\
5376.5 57436.469645693345 \\
5394.5 58115.58285381549 \\
5458.0 60507.520411178324 \\
5899.0 76954.89623846291 \\
6126.0 85308.99316673828 \\
6128.5 85400.57481375406 \\
6228.5 89056.26907945558 \\
6454.0 97245.63931409596 \\
6472.0 97896.09857654496 \\
6530.5 100006.78533855622 \\
6570.5 101447.07950724382 \\
6609.5 102849.09040507034 \\
6624.0 103369.77782510396 \\
6660.5 104679.09866535827 \\
6750.5 107899.15097841201 \\
7049.5 118510.96207549208 \\
7189.0 123416.77107935882 \\
7327.5 128258.9710531371 \\
7365.0 129565.16064666584 \\
7599.0 137668.85373290512 \\
7633.0 138839.58243110275 \\
7884.0 147429.47239428887 \\
7929.5 148976.63830383553 \\
7943.0 149435.0991598629 \\
8727.5 175614.3234736853 \\
8912.5 181655.39145447424 \\
8984.5 183992.841007396 \\
9001.5 184543.62117426936 \\
9125.5 188548.1611807748 \\
9161.5 189706.5146915164 \\
9164.0 189786.88480822922 \\
9282.5 193585.8365744957 \\
9401.5 197379.94032999023 \\
9446.5 198809.23388486437 \\
9740.0 208058.0149924528 \\
10060.0 217996.84448888554 \\
10264.5 224269.14519586804 \\
10927.5 244179.41969217645 \\
10940.0 244548.56459271192 \\
10940.0 244548.56459271192 \\
11889.0 271900.01400136005 \\
13599.0 317825.74034848285 \\
13664.0 319486.22825632134 \\
    };
\end{axis}
\end{tikzpicture}
\end{tabular}
}
\caption{The \textbf{runtime and memory consumption} on high resolution shapes highly relate on the problem instance and the feature quality. The green line visualizes the trend.}
\label{fig:runtimeAndMemory}
\end{figure}
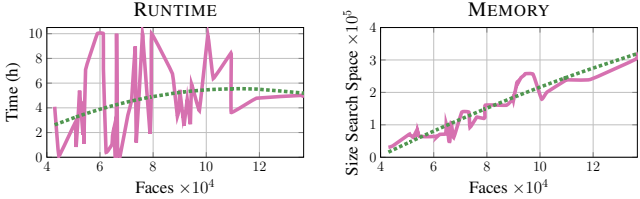

Our method's scalability is constrained by quadratically growing memory consumption and exponentially increasing runtime.
However, our coarse-to-fine scheme extends our method's applicability to high-resolution shapes by propagating the matching to the full resolution~(in the main paper, we use the full resolution given in the datasets).
In Figure~\ref{fig:runtimeAndMemory} we show runtime and memory consumption (w.r.t.~search space size) of high resolution shapes. Note that the runtime and memory consumption heavily rely on the problem instance.
The green line shows the trend of runtime and memory.

\subsection{Runtime vs. Performance}
\begin{figure}
\centering
\begin{minipage}{0.5\linewidth}
\newcommand{\pckLineWidth}{3pt}
\newcommand{\plotWidth}{1.2*\linewidth}
\newcommand{\plotHeight}{0.75*\linewidth}
\newcommand{\pckTitle}{\textsc{Mean IoU}/$\text{AUC}_{\text{GeoErr}}$}

\pgfplotsset{%
    label style = {font=\scriptsize},
    tick label style = {font=\tiny},
    title style =  {font=\scriptsize,
                    yshift=-0.2cm
                    },
    legend style={  fill= gray!10,
                    fill opacity=0.6, 
                    font=\tiny,
                    draw=gray!20, 
                    text opacity=1},
}
\tikzset{mark options={mark size=3, line width=1pt}}

\begin{tikzpicture}[scale=1, transform shape]
\begin{axis}[%
width=\plotWidth,
inner sep=1pt,
outer sep=0pt,
label style={inner sep=0pt},
tick label style={inner sep=1pt},
		height=\plotHeight,
		grid=major,
		title=\pckTitle,
		legend style={
			at={(0.98, 0.02)},
			anchor=south east,
			legend columns=1},
		legend cell align={left},
    xlabel={\scriptsize log(Time (s))},
    xlabel near ticks,
    ylabel={\tiny $\bigcirc$ mIoU, $\Box\;\text{AUC}_{\text{GeoErr}}$},
    ylabel near ticks,
    ymin=30,
    ymax=80,
scatter/classes={%
    DPFM={%
    fill=cPLOTdpfm,
    draw=cPLOTdpfm}, 
    SmComb={%
    fill=cPLOTroetzer,
    draw=cPLOTroetzer},
    Ours={%
    fill=cPLOTours,
    draw=cPLOTours}
}]
\addplot+[scatter,
only marks,%
    scatter src=explicit symbolic,%
    mark=o,%
visualization depends on={value \thisrow{label} \as \Label}]%
table[meta=label] {
x y label
0.18 54.93 DPFM
3.8 57.86 SmComb
9.79 69.29 Ours
    };
    \addplot[scatter,
    only marks,%
    mark=square,%
    scatter src=explicit symbolic, 
visualization depends on={value \thisrow{label} \as \Label}]%
table[meta=label] {
x y label
0.18 36.7 DPFM
3.8 52.3 SmComb
9.79 61.25 Ours
    };
\legend{DPFM,Sm-Comb,Ours}
\end{axis}

\end{tikzpicture} 
\end{minipage}
\caption{Our method exceeds the comparison methods in terms of mean IoU~(circle) and $\text{AUC}_{\text{GeoErr}}$~(square), but is slower in terms of optimization time.}
\label{fig:mIoUandAUC}
\end{figure}
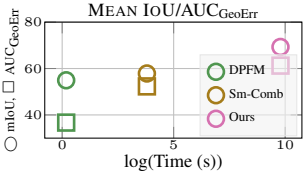

We plot mean IoU ($\uparrow$) and $\text{AUC}_{\text{GeoErr}}$($\uparrow$) over log of optimization time (see Figure~\ref{fig:mIoUandAUC}). Despite slower average opt.~time (Sm-Comb: 49s; DPFM train: 118.2min, test: 1.31s; Ours 299.05min) our method yields improved performance. We
hope that our work will inspire follow-up works that explore  improved solvers for such problems.
While our work is not competitive to current SOTA methods w.r.t.~runtime, we consider our contribution as a vital first step to solving the extremely challenging and yet underexplored partial-to-partial shape matching setting.

\subsubsection{Objective Function}

We normalize our objective function by the number of elements in $\mathbf{\Gamma}$:  $\frac{\bigl\langle\mathbf{C},\mathbf{\Gamma}\bigr\rangle}{\mathbf{1^T}\mathbf{\Gamma}}$.

We show in our ablation study in Table~\ref{tab:sum_comparison} 
that this normalization improves performance over using only the sum as the objective function $\bigl\langle\mathbf{C},\mathbf{\Gamma}\bigr\rangle$ or adding a penalty function to the sum objective $\bigl\langle\mathbf{C},\mathbf{\Gamma}\bigr\rangle - \lambda \mathbf{1^T}\mathbf{\Gamma}$ in terms of mean IoU.

\begin{table}
\begin{tabular}{c | c | c | c | c}
Sum $[0]$&Sum $[0.01]$&Sum $[0.1]$&Sum $[1]$ & \textbf{Ours}\\
\hline
{51.23}
& 
{52.53}
& 
{53.12}
& 
{52.53}
& 
{\textbf{69.29}}
\end{tabular}
\caption{
We compare \textbf{mean IoU} of our normalized objective function with the sum objective and weighting factor $\lambda$ (in brackets) on the CP2P TEST dataset. 
Our normalized version performs best.
}
\label{tab:sum_comparison}
\end{table}

\subsubsection{Performance in terms of Percentage of overlapping region}
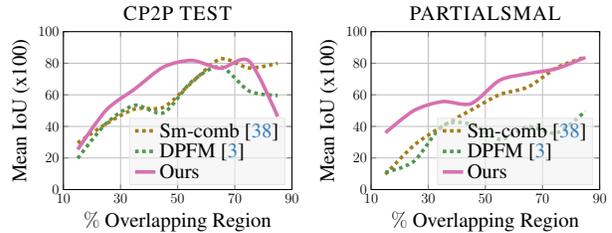
\begin{figure}[htpb]%
    \setlength\tabcolsep{-5pt} %
    \begin{tabular}{ll}
      {\newcommand{\pckLineWidth}{3pt}
\newcommand{\plotWidth}{1*\linewidth}
\newcommand{\plotHeight}{0.75*\linewidth}
\newcommand{\pckTitle}{\textsc{CP2P TEST}}

\pgfplotsset{%
    label style = {font=\LARGE},
    tick label style = {font=\large},
    title style =  {font=\LARGE},
    legend style={  fill= gray!10,
                    fill opacity=0.6, 
                    font=\LARGE,
                    draw=gray!20, 
                    text opacity=1}
}
\begin{tikzpicture}[scale=0.45, transform shape]
	\begin{axis}[
		width=\plotWidth,
		height=\plotHeight,
		grid=major,
		title=\pckTitle,
		legend style={
			at={(0.97,0.03)},
			anchor=south east,
			legend columns=1},
		legend cell align={left},
	ylabel={{\LARGE
        Mean IoU (x100)}},
        xmin=10,
        xmax=90,
        xlabel=
        $\%$ Overlapping Region,
        ylabel near ticks,
        xtick={10,30,50,70,90},
        xticklabels={$10$, $30$, $50$, $70$, $90$},
        ymin=0,
        ymax=100,
        ytick={0, 20, 40, 60, 80, 100},
        yticklabels={$0$, $20$, $40$, $60$, $80$, $100$},
	]
 \addplot [color=cPLOTroetzer, dashed, smooth, line width=\pckLineWidth]
    table[row sep=crcr]{%
15  29.50652712119412\\
25  41.662064511180304\\
35  51.03963241002931\\
45  52.13067481972268\\
55  67.87946390100214\\
65  82.71568555223588\\
75  77.02385045362851\\
85  79.99690371903526\\
    };
    \addlegendentry{\textcolor{black}{Sm-comb~\cite{roetzer2022scalable}}}
    \addplot [color=cPLOTdpfm, dashed, smooth, line width=\pckLineWidth]
    table[row sep=crcr]{%
15  19.758303680767618\\
25  41.89012087881565\\
35  53.34587062106413\\
45  48.62774527416779\\
55  68.49846847355366\\
65  77.81210601329803\\
75  62.402707835038505\\
85  59.58449443181356\\
    };
    \addlegendentry{\textcolor{black}{DPFM~\cite{attaiki2021dpfm}}}
    \addplot [color=cPLOTours, smooth, line width=\pckLineWidth]
    table[row sep=crcr]{%
15  25.365307349471884\\
25  50.35927353312603\\
35  63.71001711864987\\
45  77.59067255615099\\
55  81.74632653435803\\
65  76.80499650757142\\
75  81.35409104245093\\
85  46.20842543476183\\
    };
    \addlegendentry{\textcolor{black}{Ours}}
	\end{axis}
\end{tikzpicture}}   & \newcommand{\pckLineWidth}{3pt}
\newcommand{\plotWidth}{1*\linewidth}
\newcommand{\plotHeight}{0.75*\linewidth}
\newcommand{\pckTitle}{\textsc{PARTIALSMAL}}

\pgfplotsset{%
    label style = {font=\LARGE},
    tick label style = {font=\large},
    title style =  {font=\LARGE},
    legend style={  fill= gray!10,
                    fill opacity=0.6, 
                    font=\LARGE,
                    draw=gray!20, 
                    text opacity=1}
}
\begin{tikzpicture}[scale=0.45, transform shape]
	\begin{axis}[
		width=\plotWidth,
		height=\plotHeight,
		grid=major,
		title=\pckTitle,
		legend style={
			at={(0.97,0.03)},
			anchor=south east,
			legend columns=1},
		legend cell align={left},
	ylabel={{\LARGE
        Mean IoU (x100)}},
        xmin=10,
        xmax=90,
        xlabel=$\%$ Overlapping Region,
        ylabel near ticks,
        xtick={10,30,50,70,90},
        xticklabels={$10$, $30$, $50$, $70$, $90$},
        ymin=0,
        ymax=100,
        ytick={0, 20, 40, 60, 80, 100},
        yticklabels={$0$, $20$, $40$, $60$, $80$, $100$},
	]
 \addplot [color=cPLOTroetzer, dashed, smooth, line width=\pckLineWidth]
    table[row sep=crcr]{%
15  9.669537912546874\\
25  28.000582277581206\\
35  40.18830288990277\\
45  50.33518591407024\\
55  60.10353574329882\\
65  65.03697292527531\\
75  77.21051079564543\\
85  83.8894309291855\\
    };
    \addlegendentry{\textcolor{black}{Sm-comb~\cite{roetzer2022scalable}}}
    \addplot [color=cPLOTdpfm, dashed, smooth, line width=\pckLineWidth]
    table[row sep=crcr]{%
15  10.911873169243336\\
25  17.93290339410305\\
35  40.35381339490414\\
45  41.448278921215156\\
55  31.840045364132447\\
65  39.86905595908563\\
75  35.87482012808323\\
85  49.700612517503586\\
    };
    \addlegendentry{\textcolor{black}{DPFM~\cite{attaiki2021dpfm}}}
    \addplot [color=cPLOTours, smooth, line width=\pckLineWidth]
    table[row sep=crcr]{%
15  35.9230529844367\\
25  50.06622478971462\\
35  55.715565338850915\\
45  54.41201973560236\\
55  69.20298788091287\\
65  73.41870296642156\\
75  76.6409841038679\\
85  83.68152709099358\\
    };
    \addlegendentry{\textcolor{black}{Ours}}
	\end{axis}
\end{tikzpicture}
    \end{tabular}
\caption{
Our methods' performance increases with an \textbf{increasing percentage of overlapping region}. 
With larger overlapping regions, it performs similarly to Sm-comb~\cite{roetzer2022scalable} and DPFM~\cite{attaiki2021dpfm}. With smaller overlapping regions, our method outperforms the others.}
\label{fig:percentages}
\end{figure}

With an increasing percentage of the overlapping region, we see increased performance in our method.
For a substantial amount of overlapping region, we observe similar results between the method from Sm-comb~\cite{roetzer2022scalable}, DPFM~\cite{attaiki2021dpfm}, and our method, which makes sense as it is close to a full-to-full matching. 
For more partiality, our approach outperforms the comparison methods. 
We show the results in Figure~\ref{fig:percentages}.

\section{Failure Cases \& Artefacts}
We can not guarantee geometric consistency at boundary edges because of the lack of adjacent neighboring triangles. The interior of two triangle patches is still matched geometrically consistent, see Sec.~5 in the main paper.
We show some failure cases in Figure~\ref{fig:failure_cases}.
\def\failurerebutHEIGHT{0.182\linewidth}
\begin{figure}
\centering
\begin{tabular}{c|c|c}
\hspace{0.2cm}
\includegraphics[height=\failurerebutHEIGHT]{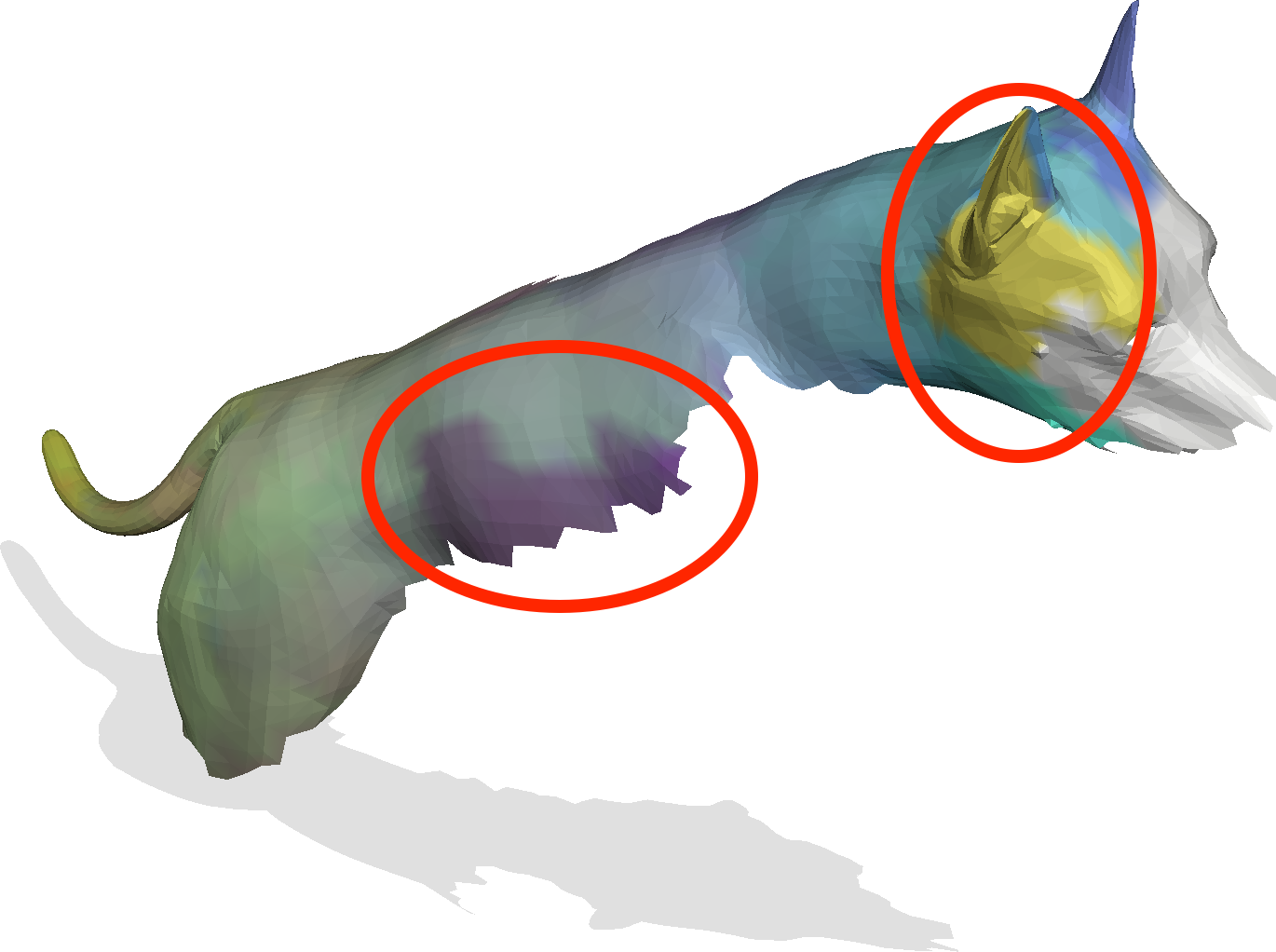}&
\includegraphics[height=\failurerebutHEIGHT]{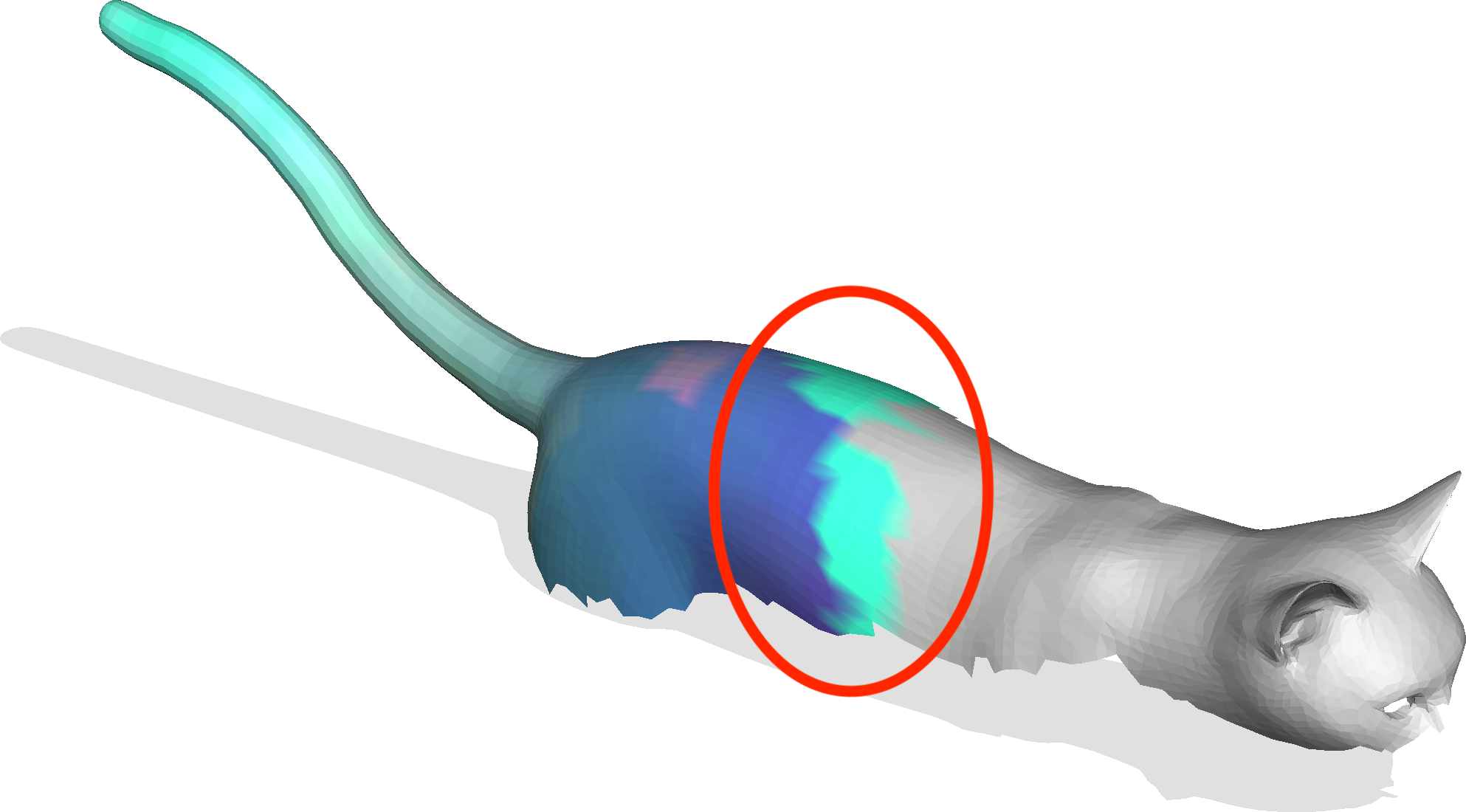}&
\includegraphics[height=\failurerebutHEIGHT]{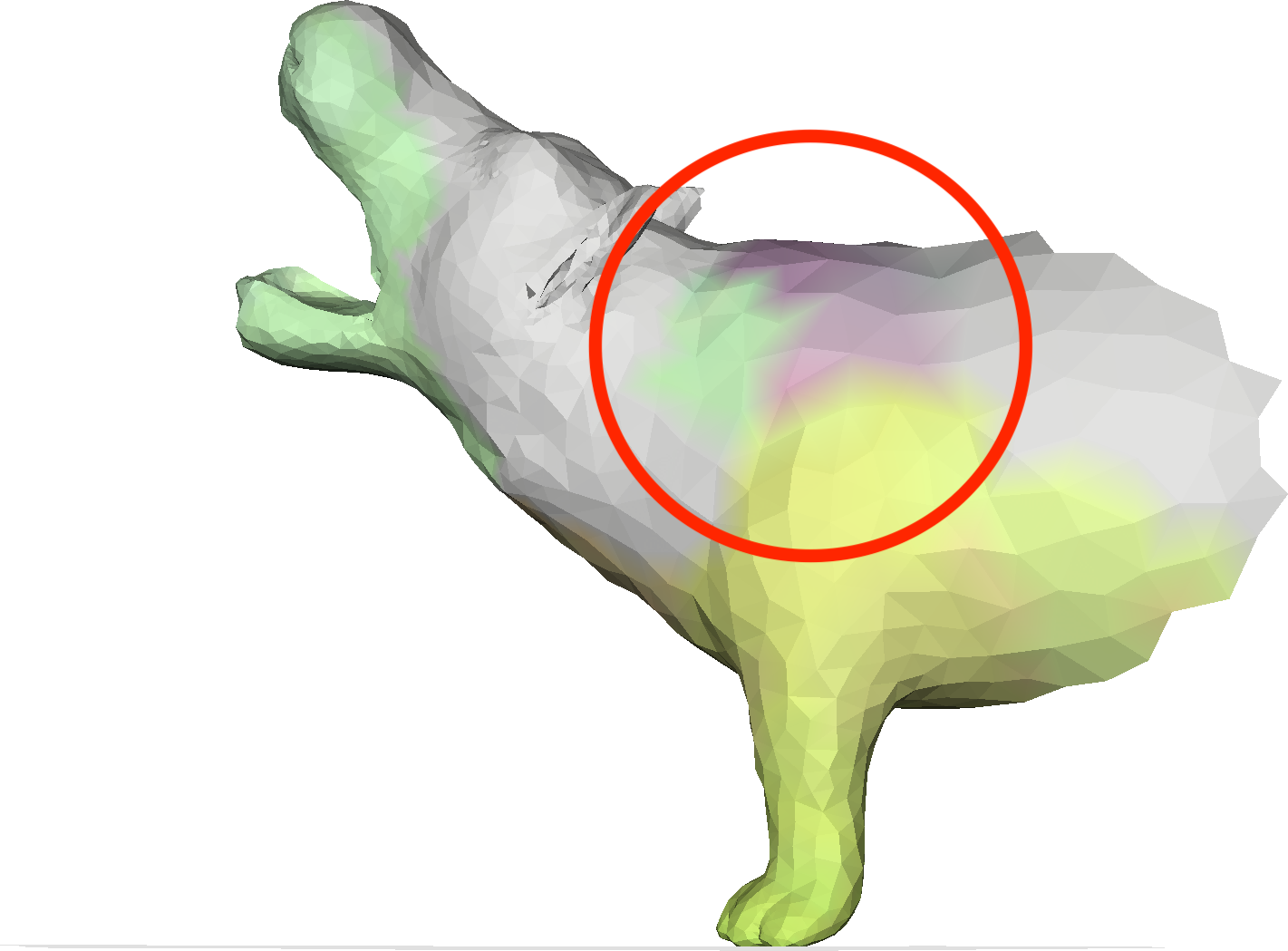}
\end{tabular}
    \caption{We observe some \textbf{failure cases} based on the missing geometric consistency on boundary edges.}
    \label{fig:failure_cases}
\end{figure}

\section{Dataset Analysis}
{
During our research we noticed that qualitatively similar shapes in the SHREC16 CUTS test dataset do also appear in the SHREC16 train dataset, although the vertex positions have some slight jitter.
In Figure~\ref{fig:dataset_analysis} we visualize two plots that show the cumulative number of test shapes (vertical axis) versus the distance between each test shape and the nearest training shape. 
While in the CUTS dataset we oberserve 47 shapes with this problem the HOLES dataset is not affected. 
\begin{figure}[h]
{
\setlength\tabcolsep{-5pt} %
\begin{tabular}{cc}
\hspace{-0.5cm}
\newcommand{\pckLineWidth}{3pt}
\newcommand{\plotWidth}{1.1*\linewidth}
\newcommand{\plotHeight}{0.65*\linewidth}
\newcommand{\pckTitle}{\textsc{Cuts}}

\pgfplotsset{%
    label style = {font=\LARGE},
    tick label style = {font=\large},
    title style =  {font=\LARGE},
    legend style={  fill= gray!10,
                    fill opacity=0.6,
                    font=\LARGE,
                    draw=gray!20, 
                    text opacity=1},
}
\begin{tikzpicture}[scale=0.45, transform shape]

\begin{axis}[
width=\plotWidth,
height=\plotHeight,
title=\pckTitle,
scaled x ticks=false,
xticklabel=\pgfkeys{/pgf/number format/.cd,fixed,precision=2,zerofill}\pgfmathprintnumber{\tick},
tick align=outside,
tick pos=left,
max space between ticks=50,
xmajorgrids,
xmin=-0.019921817, xmax=0.418358157,
xlabel={Mean Min Dist},
ymajorgrids,
ymin=-9.95, ymax=208.95,
ylabel={Num Shapes $<$ Mean Min Dist}
]
\addplot [color=cPLOTours, smooth, tension=0.1, line width=\pckLineWidth]
table {%
        0 0
        0.00200219266331658 44
        0.00400438532663317 47
        0.00600657798994975 47
        0.00800877065326633 47
        0.0100109633165829 47
        0.0120131559798995 48
        0.0140153486432161 48
        0.0160175413065327 48
        0.0180197339698492 48
        0.0200219266331658 49
        0.0220241192964824 50
        0.024026311959799 50
        0.0260285046231156 51
        0.0280306972864322 51
        0.0300328899497487 53
        0.0320350826130653 53
        0.0340372752763819 54
        0.0360394679396985 57
        0.0380416606030151 57
        0.0400438532663317 62
        0.0420460459296482 65
        0.0440482385929648 65
        0.0460504312562814 65
        0.048052623919598 67
        0.0500548165829146 70
        0.0520570092462312 74
        0.0540592019095477 75
        0.0560613945728643 76
        0.0580635872361809 77
        0.0600657798994975 78
        0.0620679725628141 79
        0.0640701652261307 81
        0.0660723578894472 81
        0.0680745505527638 81
        0.0700767432160804 83
        0.072078935879397 83
        0.0740811285427136 84
        0.0760833212060301 88
        0.0780855138693467 89
        0.0800877065326633 90
        0.0820898991959799 90
        0.0840920918592965 90
        0.0860942845226131 90
        0.0880964771859297 91
        0.0900986698492462 92
        0.0921008625125628 92
        0.0941030551758794 94
        0.096105247839196 97
        0.0981074405025126 97
        0.100109633165829 97
        0.102111825829146 101
        0.104114018492462 103
        0.106116211155779 106
        0.108118403819095 106
        0.110120596482412 108
        0.112122789145729 108
        0.114124981809045 109
        0.116127174472362 110
        0.118129367135678 113
        0.120131559798995 114
        0.122133752462312 115
        0.124135945125628 115
        0.126138137788945 116
        0.128140330452261 117
        0.130142523115578 118
        0.132144715778894 121
        0.134146908442211 122
        0.136149101105528 125
        0.138151293768844 127
        0.140153486432161 127
        0.142155679095477 128
        0.144157871758794 129
        0.146160064422111 132
        0.148162257085427 132
        0.150164449748744 133
        0.15216664241206 137
        0.154168835075377 138
        0.156171027738693 139
        0.15817322040201 139
        0.160175413065327 141
        0.162177605728643 143
        0.16417979839196 143
        0.166181991055276 143
        0.168184183718593 144
        0.17018637638191 145
        0.172188569045226 145
        0.174190761708543 146
        0.176192954371859 147
        0.178195147035176 150
        0.180197339698492 152
        0.182199532361809 152
        0.184201725025126 153
        0.186203917688442 157
        0.188206110351759 157
        0.190208303015075 158
        0.192210495678392 159
        0.194212688341709 160
        0.196214881005025 162
        0.198217073668342 162
        0.200219266331658 162
        0.202221458994975 163
        0.204223651658291 164
        0.206225844321608 166
        0.208228036984925 167
        0.210230229648241 168
        0.212232422311558 169
        0.214234614974874 169
        0.216236807638191 169
        0.218239000301508 170
        0.220241192964824 170
        0.222243385628141 172
        0.224245578291457 174
        0.226247770954774 179
        0.22824996361809 179
        0.230252156281407 179
        0.232254348944724 181
        0.23425654160804 181
        0.236258734271357 181
        0.238260926934673 181
        0.24026311959799 181
        0.242265312261307 183
        0.244267504924623 183
        0.24626969758794 183
        0.248271890251256 183
        0.250274082914573 183
        0.252276275577889 183
        0.254278468241206 184
        0.256280660904523 185
        0.258282853567839 185
        0.260285046231156 185
        0.262287238894472 185
        0.264289431557789 186
        0.266291624221106 186
        0.268293816884422 186
        0.270296009547739 186
        0.272298202211055 186
        0.274300394874372 186
        0.276302587537688 186
        0.278304780201005 187
        0.280306972864322 188
        0.282309165527638 189
        0.284311358190955 189
        0.286313550854271 189
        0.288315743517588 190
        0.290317936180905 190
        0.292320128844221 190
        0.294322321507538 190
        0.296324514170854 190
        0.298326706834171 190
        0.300328899497487 190
        0.302331092160804 190
        0.304333284824121 190
        0.306335477487437 191
        0.308337670150754 191
        0.31033986281407 191
        0.312342055477387 191
        0.314344248140703 191
        0.31634644080402 192
        0.318348633467337 192
        0.320350826130653 193
        0.32235301879397 193
        0.324355211457286 193
        0.326357404120603 194
        0.32835959678392 196
        0.330361789447236 196
        0.332363982110553 196
        0.334366174773869 196
        0.336368367437186 196
        0.338370560100503 196
        0.340372752763819 196
        0.342374945427136 196
        0.344377138090452 196
        0.346379330753769 197
        0.348381523417085 197
        0.350383716080402 197
        0.352385908743719 197
        0.354388101407035 197
        0.356390294070352 197
        0.358392486733668 197
        0.360394679396985 197
        0.362396872060302 197
        0.364399064723618 197
        0.366401257386935 197
        0.368403450050251 197
        0.370405642713568 197
        0.372407835376884 198
        0.374410028040201 198
        0.376412220703518 198
        0.378414413366834 198
        0.380416606030151 198
        0.382418798693467 199
        0.384420991356784 199
        0.386423184020101 199
        0.388425376683417 199
        0.390427569346734 199
        0.39242976201005 199
        0.394431954673367 199
        0.396434147336683 199
        0.39843634 199
    };
\end{axis}

\end{tikzpicture} &
\hspace{-0.1cm}
\newcommand{\pckLineWidth}{3pt}
\newcommand{\plotWidth}{1.1*\linewidth}
\newcommand{\plotHeight}{0.65*\linewidth}
\newcommand{\pckTitle}{\textsc{Holes}}

\pgfplotsset{%
    label style = {font=\LARGE},
    tick label style = {font=\large},
    title style =  {font=\LARGE},
    legend style={  fill= gray!10,
                    fill opacity=0.6,
                    font=\LARGE,
                    draw=gray!20, 
                    text opacity=1},
}
\begin{tikzpicture}[scale=0.45, transform shape]
\begin{axis}[
width=\plotWidth,
height=\plotHeight,
title=\pckTitle,
scaled x ticks=false,
xticklabel=\pgfkeys{/pgf/number format/.cd,fixed,precision=2,zerofill}\pgfmathprintnumber{\tick},
tick align=outside,
tick pos=left,
max space between ticks=50,
xmajorgrids,
xmin=-0.0321908385, xmax=0.6760076085,
xlabel={Mean Min Dist},
ymajorgrids,
ymin=-9.95, ymax=208.95,
ylabel={Num Shapes $<$ Mean Min Dist}
]
\addplot [color=cPLOTours, smooth, tension=0.1, line width=\pckLineWidth]
table {%
0 0
0.00323526015075377 14
0.00647052030150754 19
0.00970578045226131 25
0.0129410406030151 27
0.0161763007537688 32
0.0194115609045226 38
0.0226468210552764 45
0.0258820812060302 48
0.0291173413567839 55
0.0323526015075377 63
0.0355878616582915 67
0.0388231218090452 71
0.042058381959799 71
0.0452936421105528 73
0.0485289022613065 74
0.0517641624120603 77
0.0549994225628141 79
0.0582346827135678 81
0.0614699428643216 84
0.0647052030150754 86
0.0679404631658291 90
0.0711757233165829 96
0.0744109834673367 102
0.0776462436180905 104
0.0808815037688442 105
0.084116763919598 108
0.0873520240703518 115
0.0905872842211055 117
0.0938225443718593 120
0.0970578045226131 121
0.100293064673367 123
0.103528324824121 125
0.106763584974874 129
0.109998845125628 133
0.113234105276382 137
0.116469365427136 138
0.119704625577889 142
0.122939885728643 143
0.126175145879397 143
0.129410406030151 145
0.132645666180905 149
0.135880926331658 150
0.139116186482412 150
0.142351446633166 153
0.14558670678392 155
0.148821966934673 156
0.152057227085427 157
0.155292487236181 159
0.158527747386935 159
0.161763007537688 159
0.164998267688442 161
0.168233527839196 165
0.17146878798995 166
0.174704048140704 167
0.177939308291457 167
0.181174568442211 168
0.184409828592965 169
0.187645088743719 170
0.190880348894472 172
0.194115609045226 172
0.19735086919598 172
0.200586129346734 172
0.203821389497487 172
0.207056649648241 172
0.210291909798995 174
0.213527169949749 175
0.216762430100503 177
0.219997690251256 178
0.22323295040201 178
0.226468210552764 179
0.229703470703518 180
0.232938730854271 181
0.236173991005025 182
0.239409251155779 183
0.242644511306533 185
0.245879771457286 187
0.24911503160804 188
0.252350291758794 188
0.255585551909548 189
0.258820812060301 189
0.262056072211055 189
0.265291332361809 190
0.268526592512563 191
0.271761852663317 191
0.27499711281407 192
0.278232372964824 192
0.281467633115578 192
0.284702893266332 192
0.287938153417085 192
0.291173413567839 192
0.294408673718593 194
0.297643933869347 194
0.300879194020101 194
0.304114454170854 194
0.307349714321608 194
0.310584974472362 194
0.313820234623116 194
0.317055494773869 194
0.320290754924623 195
0.323526015075377 195
0.326761275226131 195
0.329996535376884 195
0.333231795527638 195
0.336467055678392 195
0.339702315829146 195
0.3429375759799 195
0.346172836130653 196
0.349408096281407 196
0.352643356432161 196
0.355878616582915 196
0.359113876733668 196
0.362349136884422 197
0.365584397035176 197
0.36881965718593 197
0.372054917336683 197
0.375290177487437 197
0.378525437638191 197
0.381760697788945 198
0.384995957939699 198
0.388231218090452 198
0.391466478241206 198
0.39470173839196 199
0.397936998542714 199
0.401172258693467 199
0.404407518844221 199
0.407642778994975 199
0.410878039145729 199
0.414113299296482 199
0.417348559447236 199
0.42058381959799 199
0.423819079748744 199
0.427054339899498 199
0.430289600050251 199
0.433524860201005 199
0.436760120351759 199
0.439995380502513 199
0.443230640653266 199
0.44646590080402 199
0.449701160954774 199
0.452936421105528 199
0.456171681256281 199
0.459406941407035 199
0.462642201557789 199
0.465877461708543 199
0.469112721859297 199
0.47234798201005 199
0.475583242160804 199
0.478818502311558 199
0.482053762462312 199
0.485289022613065 199
0.488524282763819 199
0.491759542914573 199
0.494994803065327 199
0.49823006321608 199
0.501465323366834 199
0.504700583517588 199
0.507935843668342 199
0.511171103819095 199
0.514406363969849 199
0.517641624120603 199
0.520876884271357 199
0.524112144422111 199
0.527347404572864 199
0.530582664723618 199
0.533817924874372 199
0.537053185025126 199
0.540288445175879 199
0.543523705326633 199
0.546758965477387 199
0.549994225628141 199
0.553229485778895 199
0.556464745929648 199
0.559700006080402 199
0.562935266231156 199
0.56617052638191 199
0.569405786532663 199
0.572641046683417 199
0.575876306834171 199
0.579111566984925 199
0.582346827135678 199
0.585582087286432 199
0.588817347437186 199
0.59205260758794 199
0.595287867738693 199
0.598523127889447 199
0.601758388040201 199
0.604993648190955 199
0.608228908341709 199
0.611464168492462 199
0.614699428643216 199
0.61793468879397 199
0.621169948944724 199
0.624405209095477 199
0.627640469246231 199
0.630875729396985 199
0.634110989547739 199
0.637346249698493 199
0.640581509849246 199
0.64381677 199
};
\end{axis}
\end{tikzpicture}\end{tabular}
}
\caption{The Mean Minimum distance of the SHREC16 dataset is almost zero for 47 shape in the SHREC16 CUTS dataset. 
The SHREC16 HOLES dataset is not affected by this problem.
}
\label{fig:dataset_analysis}
\end{figure}
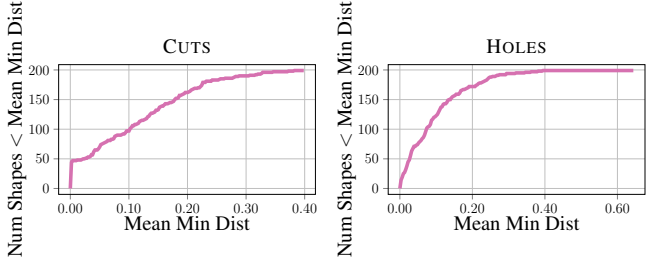
}

\section{Search Space Pruning}
Below we elaborate on the bounds that we use for our search space pruning.
Let $j, k \in \mathbb{N}$.
We know that $g(j+1) - g(j) \leq 1$. We start from this and first derive that $h(j+1) - h(j) \leq \frac{1}{j}$, where $h(j) = \frac{g(j)}{j}$.

Starting with
\begin{align}
    g(j+1) - g(j) \leq 1,\\
    \intertext{we divide by $j$, so that we get}
     \frac{g(j+1)}{j} - \frac{g(j)}{j} \leq \frac{1}{j}.\\
     \intertext{Plugging in $h(j) = \frac{g(j)}{j}$ gives}
     \frac{g(j+1)}{j} - h(j) \leq \frac{1}{j}.\\
     \intertext{As $g(j+1)\geq 0$ and $j+1\geq0$,  $\frac{g(j+1)}{j} \geq \frac{g(j+1)}{j+1}$, so that we get}
     \frac{g(j+1)}{j+1} - h(j) \leq \frac{1}{j}.\\
     \intertext{By definition $h(j+1) = \frac{g(j+1)}{j+1}$, so that we obtain}
     h(j+1) - h(j) \leq \frac{1}{j}. \label{eq:ineq}
\end{align}
Let us now consider $h(k) - h(j)$, where $k \geq j$. We can use inequality \eqref{eq:ineq} to bound this expression from above: 
We know that $h(j+1) - h(j) \leq \frac{1}{j}$,  $h(j+2) - h(j+1) \leq \frac{1}{j+1}$, .., $h(k) - h(k-1) \leq \frac{1}{k-1}$. By adding up this sequence of inequalities we obtain
$h(k)-h(j)\leq \sum_{i=j}^{k-1} \frac{1}{i}$.

\end{document}